\documentclass{article}

\usepackage{arxiv}
\usepackage[utf8]{inputenc} 
\usepackage[T1]{fontenc}    
\usepackage{hyperref}       
\usepackage{url}            
\usepackage{booktabs}       
\usepackage{amssymb,amsfonts}       
\usepackage{nicefrac}       
\usepackage{microtype}      
\usepackage{lipsum}
\usepackage{graphicx}
\usepackage{multirow}
\usepackage{algorithmic}
\usepackage{algorithm}
\usepackage{color,array}
\usepackage{times}
\usepackage{tabularx}
\usepackage{graphicx}
\usepackage{multicol}
\usepackage{url}
\usepackage{subfig}
\usepackage{booktabs}    
\makeatletter
\newcommand\urlfootnote@[1]{\footnote{\url@{#1}}}
\DeclareRobustCommand{\urlfootnote}{\hyper@normalise\urlfootnote@}
\makeatother

\title{Negative Selection Algorithm Research and Applications in the last decade: A Review }
\author{
 Kishor Datta Gupta \\
  University of Memphis\\
  Tennessee, USA \\
  \texttt{kgupta1@memphis.edu} \\
  
  \And

 Dipankar Dasgupta \\
   University of Memphis\\
  Tennessee, USA \\
  \texttt{ddasgupt@memphis.edu} \\
}

\begin{document}
\maketitle
\begin{abstract}
The Negative selection Algorithm (NSA) is one of the important methods in the field of Immunological Computation (or Artificial Immune Systems). Over the years, some progress was made which turns this algorithm (NSA) into an efficient approach to solve problems in different domain. This review takes into account these signs of progress during the last decade and categorizes those based on different characteristics and performances. Our study shows that NSA’s evolution can be labeled in four ways highlighting the most notable NSA variations and their limitations in different application domains. We also present alternative approaches to NSA for comparison and analysis. It is evident that NSA performs better for nonlinear representation than most of the other methods, and it can outperform neural-based models in computation time. We summarize NSA’s development and highlight challenges in NSA research in comparison with other similar models. 
\end{abstract}


\section{Introduction}
Negative Selection Algorithm (NSA) is a bio-inspired approach and a powerful computational tool for some applications where the decision lies in the complementary space of positive profile (data). Existing works showed that NSAs are beneficial for one class classifications, outlier detection, fault and intrusion detection problems. Inspired by the human immune system, NSA variations emerged with different representation, distance measures, coverage estimations for different applications over last three decades.  We briefly reviewed these and compared NSA with other ML models such as neural network-based techniques, discussed their advantages and limitations with directions in future research. NSA method is most compared with one class support vector machine (OCSVM). OCSVM\cite{chen2001one} and NSA both can work with one class of data. But NSA can work with high dimension data where OCSVM performance fail when data start to became non linear. Some similar other known approaches are Isolation forest (IF) \cite{liu2008isolation}, Local outlier factors (LOF) \cite{breunig2000lof}, Minimum Co-variance Method (MCM) \cite{hubert2018minimum}, Gaussian Mixture Model (GMM)\cite{reynolds2009gaussian}, Dirichlet Process Mixture Model (DPMM)\cite{blei2006variational}, Kernel Density Estimator (KDE)\cite{moon1995estimation}, Robust KDE\cite{kim2012robust}. GWR-Netwrok\cite{marsland2002self}, Deep Support Vector Data Description (SVDD), Angle based Outlier Detection (ABOD)\cite{kriegel2008angle}, Subspace Outlier Detection (SOD) \cite{kriegel2009outlier}, Deep Auto Encoder based Methods\cite{hinton2006reducing}, Generative Adversarial Net Based approaches (eg: \cite{li2018anomaly,Schlegl2017Unsupervised}. In this paper, we compared these methods with NSA approaches for outlier detection data-sets and also used a synthetic data set to visualize the comparison.
 \begin{figure}
\centering
    \includegraphics[width=1\linewidth]{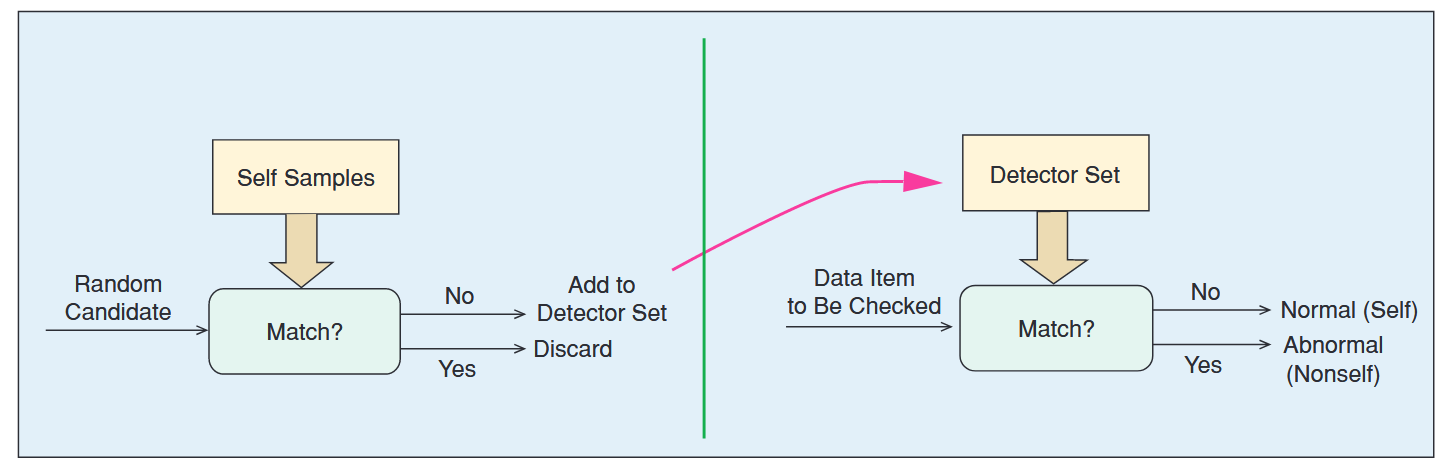}
        \caption{The basic Negative Selection Algorithm (NSA) \cite{dasgupta2006advances} similar to any two-phase supervised learning algorithms. The left diagram shows detector generation in the complementary space (training phase) and the right illustrates the use of detectors (testing phase). }
        \label{fig:bas}
\end{figure}

Contributions of this survey paper are:
\begin{itemize}
    \item A brief study of NSA research summary for last 30 years.
    \item Applications of NSA in the last decade.
    \item A visual comparison of NSA with similar algorithms.
\end{itemize}
Following this section, we give a short overview of NSA and related review works. In section 3, we discuss different variants of NSA; NSA applications are discussed in the section 4. The section 5 discusses the alternatives for NSA with some empirical comparisons. 
 
\section{Immunological Computation:} 
Immunological Computation a.k.a. Artificial Immune System (AIS) is inspired by the human immune system (HIS) mechanism and utilizes to solve computational problems\cite{dasgupta2003artificial}. One of the fundamental aspects of the HIS is self/non-self discrimination. The Human immune system can identify which cells are own (self) and can differentiate foreign entities (non-self)\cite{von1990self}. Therefore, it can strengthen its defense versus the adversarial rather of hurting the self cell. The most popular AIS research methods include the Negative Selection Algorithm (NSA), clonal selection, immune network theory, danger theory, and positive selection \cite{dasgupta2008immunological}.

\begin{figure}
\centering
\subfloat[Two dimensional projections of data points (self profile) and negative detectors in representation space in NSA\cite{ji2007revisiting} ]{
        \includegraphics[width=0.44\linewidth]{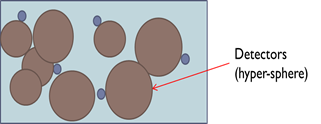}
   }
    \subfloat[ A sample representation of the grid-based NSA\cite{yang2010gf} ]{
        \includegraphics[width=0.44\linewidth]{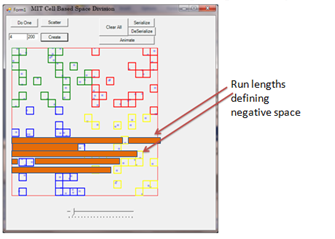}  
    }
     \caption{Two different representation of NSA}
        \label{fig:nd}
\end{figure}
\subsection{Basic concepts of NSA}
 The NSA is one of the most studied and researched algorithms, particularly for anomaly detection\cite{ji2007revisiting}. In 1994 \cite{forrest1994self}, introduced NSA for solving computer security problems.
  \begin{figure}
\centering
    \includegraphics[width=0.65\linewidth]{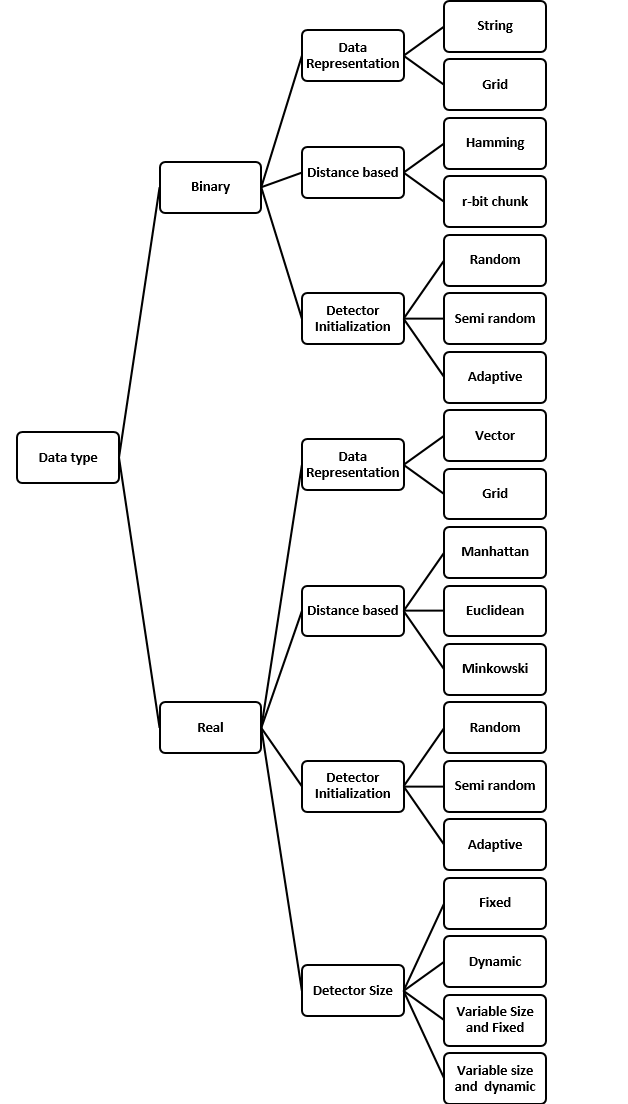}
        \caption{Classifications of NSA}
        \label{fig:class}
\end{figure}
As described in figure \ref{fig:bas},in traditional NSA implementations, detectors are generated randomly in the representation space. If these detectors are not matched with self (positive) data, they are stored as negative detectors. These detectors later use to classify data from self and non-self. Figure \ref{fig:nd} illustrates two different representation of NSA, in particular, Figure \ref{fig:nd}(a) real-valued V-detector based NSA and Figure \ref{fig:nd}(b) Grid-Based NSA, detailed discussion of these approaches covered in a later section.

\subsubsection{NSA Terminology}
\begin{itemize}
    \item Self: Representation of a data class. NSA will identify a given data is from self class or not. This self data can be a set of real values or a set binary value, or a string.
    \item Detector: A set of data which matched with non-self data.
    \item Distance Measure/Matching rules: The formula/method used to measure the distance between two data points in representation space. Commonly in the NSA, it was used to measure the detector's distance from a data point. It is also known as matching rules, primarily when data is represented in string representation. Examples are Euclidean distance, Manhattan distance, R-bit chunk matching, hamming Distance, etc.
\end{itemize}
Most of these terminology and other necessary mathematics related to distance measures have been detailed by the \cite{ji2007revisiting} with a statistical explanation. 

 Based on feature value, we can have two kinds of NSA: the binary NSA (BNSA) and the Real-Value NSA (RNSA). Real-value NSA can be variable size or constant size. The BNSA use r-contiguous bits(rcb), r-chunks, landscape-affinity matching, Hamming distance to match the similarity, where RNSA uses mostly derivations of Euclidean distances. One example of a variable size detector or V-detector, V-detector's aim, deals with constant size detectors' drawbacks. In this algorithm, the size radius of detectors is changing from one to the others\cite{ji2009v}. 
 In figure \ref{fig:class}, we illustrated the different classifications of NSA. The NSA detectors can be represented by string for (binary) BNSA or by a vector in multidimensional space for real values. Later, grid-based representations were also introduced by Yang\cite{yang2010gf}. A variation of grid representation \cite{4630915} is known as matrix representation also introduced as an additional approach. Detector initialization was random in most of the early variations of NSA. Later, researchers tried some heuristic methods to use pseudo-randomness and some evolutionary computation-based adaptiveness technique to initialize detector position. Detector size can be fixed or can change through the generation process. Also, the different detector can take the detector's variable size, and these sizes can be dynamically changes throughout the generation process. 

\subsection{Prior NSA Review Reports}
Dasgupta\cite{dasgupta1997immunity} first reviewed NSA as a part of immunity based system, and later \cite{dasgupta1999artificial} discusses NSA for industrial application. In 2000, another paper\cite{de2000artificial} also review NSA application as a part of artificial immune system and in 2003 other researchers\cite{dasgupta2003artificial,gonzalez2003study} also did similar study. Ji\cite{ji2007revisiting} provided a organized review of NSA. In 2014, Lasisi\cite{lasisi2014negative} did a survey on the epistemology of generating NSA detectors. Ramdane\cite{ramdane2017negative} provided a brief review of recent NSA improvements and its intrusion detection applications. Ahsan\cite{ahsan2020applications} wrote a brief report on NSA based works only for cloud security. Most of the other works were mostly outdated or focused on application specific NSA works.

\section{Evolution's of NSA}
The evolution of the NSA can be divided into four periods. In the early period, NSA was limited by string matching; researchers were focused on using different binary matching techniques. After Real Value NSA was introduced, researchers focused on detector size and position. Next was the adaptive methods to improve NSA accuracy and reduce computational cost. In the current period, research was getting more focused on combining NSA with another algorithm to customize NSA focusing specific application area.
 \begin{figure}
\centering
    \includegraphics[width=1.0\linewidth]{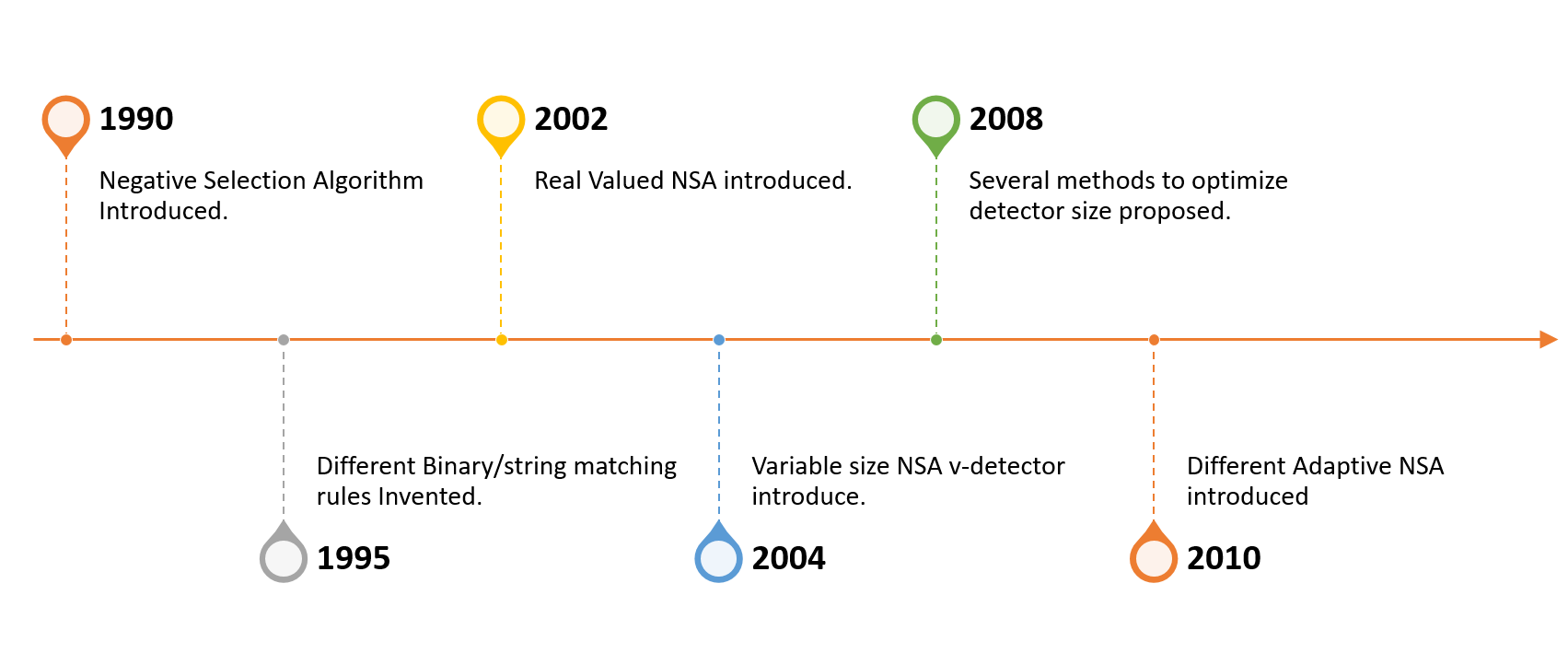}
        \caption{Major events of NSA evolution\cite{ dasgupta2003artificial,gonzalez2003study,lasisi2014negative,ramdane2017negative,ji2007revisiting}.}
        \label{fig:mnse}
\end{figure}
\subsection{String-based NSA (1986-2002)} 
After Forrest\cite{forrest1994self} introduced NSA for solving computer security problems, there were several works done by other researchers\cite{dasgupta1995tool,dasgupta1996novelty,deaton1997dna,dasgupta1999artificial,kim1999negative} but mostly these works had some common characteristics, such as:
\begin{itemize}
    \item Detectors were generated randomly.
    \item Detectors are string.
     \item Self data were encoded to string.
     \item r-contiguous bits (rcb), r-chunks, landscape-affinity matching, Hamming distance were used
\end{itemize}
These characteristics limited the applications of NSA \cite{dasgupta1997immunity}. Main reason of this limitations are generating efficient detectors are computationally costly. Helman\cite{helman1994efficient} proposed an improvement on original NSA and some other variations were proposed\cite{ayara2002negative}. But Gonzalez\cite{gonzalez2003effect} identified that binary matching rules are limited the application areas of NSA.

\subsection{Introduction and improvements of Real Value NSA (2002-2006)} 
\begin{figure}
\centering
    \includegraphics[width=1.0\linewidth]{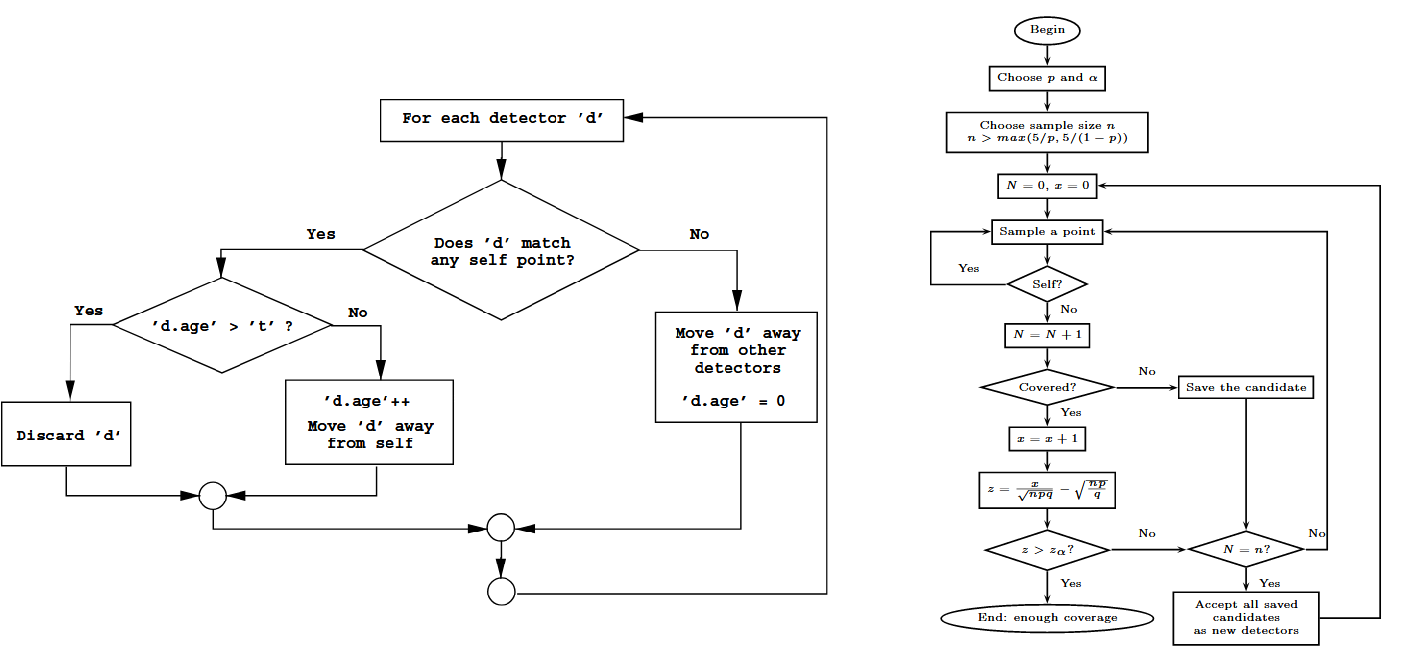}
        \caption{RNSA(left)\cite{gonzalez2003randomized} and V-detector(right)\cite{ji2005boundary} flow chart.}
        \label{fig:rnsavdet}
\end{figure}

Major breakthrough came when Gonzalez\cite{gonzalez2002combining} proposed real valued NSA. Instead of string matching this new algorithm proposed match using distances among real values. This algorithm brought so many new features in NSA due to its high level representation of data such as radius based matching, Euclidean distance, Minkowski distances etc. Real valued NSA (RNSA) quickly became popular and RNSA started to uses in many different domain and it was observed it is performing better than Binary valued NSA (BNSA)\cite{gonzalez2003anomaly}. One of the notable works for RNSA was its use in airplane fault detection\cite{dasgupta2004negative}. There were many variations of RNSA started to proposed by researchers. Gonzalez\cite{gonzalez2003randomized} Proposed RRNSA which is a Monte-Carlo based random RNSA. However most influential and popular variation of RNSA was proposed by Ji\cite{ji2004real}, known as V-detector. V-detector supports variable size of detector radius. The V-detector also get improved in subsequent works \cite{ji2005boundary} which limit the detector numbers and \cite{ji2005estimating} proposed a boundary aware method. In this period, it was noticed that NSA still has limitations like choosing the appropriate matching rules is hard and determining how to answer the dimensionality of data\cite{ji2006applicability}. We illustrated the flowchart of v-detector and RNSA in the figure \ref{fig:rnsavdet}, the flowchart shows the basic differences between v-detector and RNSA as in v-detector size of radius is not static.

One of the significant advancements in this era was the starting of negative data representation\cite{Esponda2004EnhancingPT,darlington2005negative} or negative database based research works. Negative dataset\cite{esponda2007relational} and negative data representation use the NSA. Negative database idea later boldly impacted research on Authentication and data security.

\subsection{Research towards Adaptive/Dynamic NSA(2006-2012)}
In this period, there were different approaches to improve NSA by apply different types of algorithms in various steps of NSA. However, researchers mostly focused on detector size determination and better generation technique. 
Noticeable methods was Genetic Algorithm (GA) and Fuzzy techniques in this age. Gao\cite{gao2006genetic} proposed GA based NSA, where detectors was optimized by GA. Limitations of this method is higher computational cost. A better GA based approach proposed by Bakicki\cite{balicki2006negative} where a rank based system introduced with multi objective GA but it was also application specific. Jian\cite{jian2007negative} and Wang\cite{wang2007immune} was similar GA based optimization approach with combination of fuzzy logic, but these methods also has limited applicability.  Gao\cite{gao2006clonal} uses clonal optimization and tehir subsequent works \cite{gao2007particle} uses particle optimization to determine detector size.
One of the novel approach\cite{4233937} proposes multi shape detector for better coverage, another similar work but using different shapes was proposed by Xia\cite{xia2007shape}. But both method  suffered by higher computational cost. Neural Network based approach to make NSA adaptive also proposed by Gao\cite{Gao2007ANN}. A self-adaptive NSA (ANSA) introduced an innovative method to accommodate the self range adaptively and grow the non-self-covering detectors to create a suitable system profile only utilizing a subset of self samples\cite{jinquan2009self}. Another self-adaptive NSA \cite{NSSAC} introduced attempted to make detector size and position dynamic. Some different types of the algorithm like chaos theory were applied on NSA by Zhang\cite{zhang2006negative} and a state graph supported NSA was proposed by Luo\cite{luo2007novel}.
   \begin{figure}
\centering
    \includegraphics[width=1.0\linewidth]{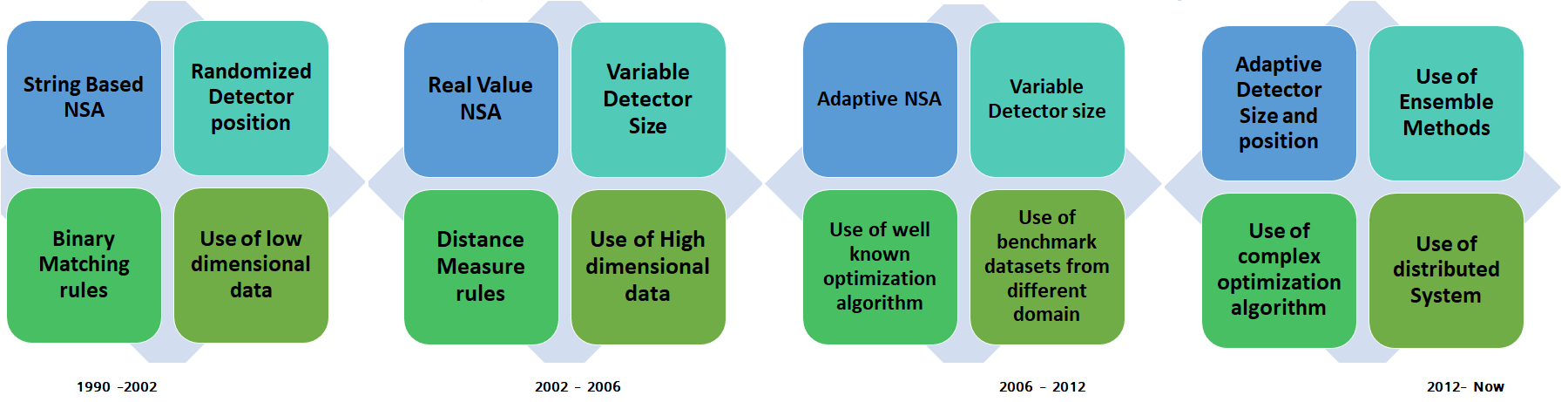}
        \caption{Common characteristics of NSA related research trend over time based on \cite{ dasgupta2003artificial,gonzalez2003study,lasisi2014negative,ramdane2017negative,ji2007revisiting}.}
        \label{fig:nste}
\end{figure}

 \subsection{Variations of Adaptive NSA (2012-now)}
 Different variations of adaptive NSA are being proposed in this period, and research also focuses on the initial detector position optimizations. It was noticeable that researchers started to add different complex algorithms to customize the NSA based on its application. Some of these variations can also be called the ensemble approach.
 Major characteristics of NSA evolved in this phase are:
 \begin{itemize}
 \item Apply adaptiveness /self learning instead of using pure random generation of detectors.
 \item Dynamic detector size generation cost was considered a major challenge and research was more focused on cost optimization.
 \item Multiple algorithm ensembles inside NSA for better optimizations.
 \end{itemize}
 
 \begin{table}
\centering
\tiny
\begin{tabular}{|l|l|l|l|l|l|}
\hline
\multicolumn{1}{|c|}{\textbf{Name}} & \multicolumn{1}{c|}{\textbf{Notes}}                                                      & \multicolumn{1}{c|}{\textbf{Type}} & \multicolumn{1}{c|}{\textbf{DM}} & \multicolumn{1}{c|}{\textbf{DS}} & \multicolumn{1}{c|}{\textbf{DI}}                                                    \\ \hline
ANSA\cite{jinquan2009self}                                & self-adaptive  NSA                                                                       & Real                               & Minkowski                                       & Dynamic                                     & Random                                                \\ \hline
EvoSeedRNSA\cite{zhang2014evoseedrnsaii,ZHANG201418}                         & use GA on initial random position   selection                                            & Real                               & Minkowski                                       & Static                                      & Adaptive                                              \\ \hline
ORNSA\cite{li2010outlier}                               & use  internal selves, boundary selves and   outlier selves                                & Real                               & Minkowski                                       & Flexible                                    & Random                                                \\ \hline
Optimized  NSA\cite{aiqiang2011optimization}                      & Optimized detector position to   reduce overlap                                           & Real                               & Minkowski                                       & Flexible                                    & Random                                                \\ \hline
FtNSA\cite{farzadnia2020novel}                               & Reduce self space                                                                        & Real                               & Minkowski                                       & Static                                      & Random                                                \\ \hline
IVRNSA\cite{wu2012improved}                              & use mature and non mature set of   detector                                              & Real                               & Minkowski                                       & Static                                      & adaptive                                              \\ \hline
CB-NSA\cite{wen2013negative}                              & hierarchical clustering used to   create selfplace                                           & Real                               & Minkowski                                       & Static                                      & semirandom                                           \\ \hline
PRR-2NSA\cite{zheng2013dual}                            & Reduce dataspace, use clustering before create detector                                & Real                               & Minkowski                                       & Static                                      & semirandom                                           \\ \hline
GF-RNSA\cite{wen2014negative}                             & Grid based representation                                                                & Real/binary                        & Minkowski/Hamming                               & Static                                      & Random                                                \\ \hline
NSA–DE\cite{idris2014hybrid}                              & Uses DE to optimize random generation of detector                                      & Real                               & Minkowski                                       & Static                                      & Random                                                \\ \hline
HNSA–IDSA\cite{ramdane2014new}                           & Use self and non self detector                                                           & Real/binary                        & Minkowski/Hamming                               & Flexible                                    & Adaptive                                              \\ \hline
NSA–PSO\cite{idris2015combined}                             & Particle Swarm optimization for random generation of detector             & Real                               & Minkowski                                       & Flexible                                    & Adaptive                                              \\ \hline
IO-RNSA\cite{xiao2015immune}                             & Morphological representation \& optimize detector position & Real                               & Minkowski                                       & Static                                      & Adaptive                                              \\ \hline
BIORV-NSA\cite{cui2015biorv}                           & self  set edge inhibition \& detector self-inhibition     & Real                               & Minkowski                                        & Dynamic                                     & Random                                                \\ \hline
NSA-II                              & Use self and non self detector                                                           & Real/binary                        & Minkowski/Hamming                               & Static                                      & Random                                                \\ \hline
OALFB-NSA\cite{li2016boundary}                 & Use two dataset to detect two set of detector                                          & Real                               & Minkowski                                       & Dynamic                                     & Adaptive                                              \\ \hline  FB-NSA\cite{li2015negativea}                  & Similar to OALFB-NSA                                        & Real                               & Minkowski                                       & Dynamic                                     & Adaptive                                              \\ \hline
DENSA\cite{wen2017parameter}                               & Statistical confidence based   detector position                                         & Real                               & Minkowski                                       & Static                                      & Adaptive                                              \\ \hline
MNSA\cite{pamukov2017multiple}                                & Uses multiple negative selection                                                         & Real                               & Minkowski                                       & Static                                      & Random                                                \\ \hline
Antigen-NSA\cite{yang2017antigen}                         & antigen space density to optimize   detector position                                     & Real                               & Minkowski                                       & Flexible                                    & Adaptive                                              \\ \hline
NSNAD\cite{guerroumi2019nsnad}                               & Optimize feature set for better   result                                                  & Real/binary                        & Minkowski/Hamming                               & Static                                      & Adaptive                                              \\ \hline
REN\cite{ren2020novel}                                 & New formulation to reduce overlap of detector coverage                                 & Real                               & Minkowski                                       & Dynamic                                     & Random                                                \\ \hline
AINSA\cite{deng2020negative}                               & Use adaptive immunology to generate detector position                                   & Real                               & Minkowski                                       & Flexible                                    & Adaptive                                              \\ \hline
ODNSA\cite{selahshoor2019intrusion}                                & Optimize cost of detector radius                                                         & Real                               & Minkowski                                       & Dynamic                                     & Adaptive                                              \\ \hline
CNSA\cite{chikh2019clustered}                                & uses clustering and fruit fly   optimization                                             & Real                               & Minkowski                                       & Dynamic                                     & Adaptive                                              \\ \hline
\end{tabular}

\caption{2012-2020 Notable NSA Models (DM: Distance Measure, DS: Detector Size,DI: Distance initialization).}
\label{tab:my-table}
\end{table}

 \begin{table*}[h]
\centering
\tiny
\begin{tabular}{|c|c|c|}
\hline
\multicolumn{2}{|c|}{Field}                     & Notes \\ \hline
\multirow{5}{*}{Computer} & Spam Detection       \cite{idris2014improved,idris2014hybrid,saleh2019intelligent,chikh2019clustered,fu2006application}        &     Clustered NSA and fruit fly optimization (CNSA–FFO) has best performance with 93\% accuracy \\ \cline{2-3} 
                          & Intrusion Detection  \cite{jin2011method,forrest1996sense,zhang2005improved,kim2002integrating,li2015negativea,7545821,dasgupta2002immunity,sim2003modeling,gonzalez2002immunogenetic,hang2004extended}       &   NSALG \cite{tosin2020negative} improved accuracy by using improved feature optimization technique \\ \cline{2-3} 
                          & Virus Detection      \cite{zhang2011malware,forrest1994self,nguyen2014combination,zeng2016extended,wu2014smartphone,zhang2016integrated,brown2016detection,lu2017ransomware,wang2012new}      &    V-NSA with mutation optimize (op-RDVD) outperforms other but not suitable for realtime protection \cite{lu2017ransomware}     \\ \cline{2-3} 
                          & Authentication         \cite{dasgupta2010password,dasgupta2014g,ngo2014largest,hassan2012negative,luo2018authentication}       & Grid based NAS provided secure authentication policy which was secure from many authentication attacks   \\ \cline{2-3} 
                          & Vision                  \cite{mahapatra2014improved,ji2006analysis,garain2006recognition,bendiab2010negative}    &  NSA can do object classification job with 90\%+ accuracy for benchmark datasets.    \\ \hline
\multicolumn{2}{|c|}{Industrial\cite{dasgupta2004negative,gao2004neural,xia2007shape,aydin2010chaotic,laurentys2010design, liu2007abnormality, surace2010novelty,gao2010multi, barontini2019deterministically,bayar2015fault,abid2017multidomain,guo2019uav,abid2020improved,jung2017combined,song2019negative,doi2020early,chen2020fault,shulin2002negative,array2002multilayered,cao2003immunogenetic,ren2006research,jia2006negative}} & For fault detection NSA\cite{ren2020novel} outperforms other NSA based approach. \\  \hline
\multicolumn{2}{|c|}{Financial \cite{ze2008artificial,butler2010modeling,hormozi2013credit,lakshminovel,mor2011rolling,wu2007artificial}  }  &     NSA has been used to detect anomaly in financial document \cite{lakshminovel}. \\ \hline
\multicolumn{2}{|c|}{Medical                 \cite{ba2012eeg,mousavi2013negative,lasisi2016application,creevey2002algorithm,perkins2019using,lee2004negative,barontini229negative,li2011hybrid} }& NSA exhibits great promises in RNA and DNA sequencing. \     \\ \hline
\end{tabular}
\caption{Notable NSA works in different domain}
\label{tab:Applications}
\end{table*}

\begin{figure}
\centering
    \includegraphics[width=1.0\linewidth]{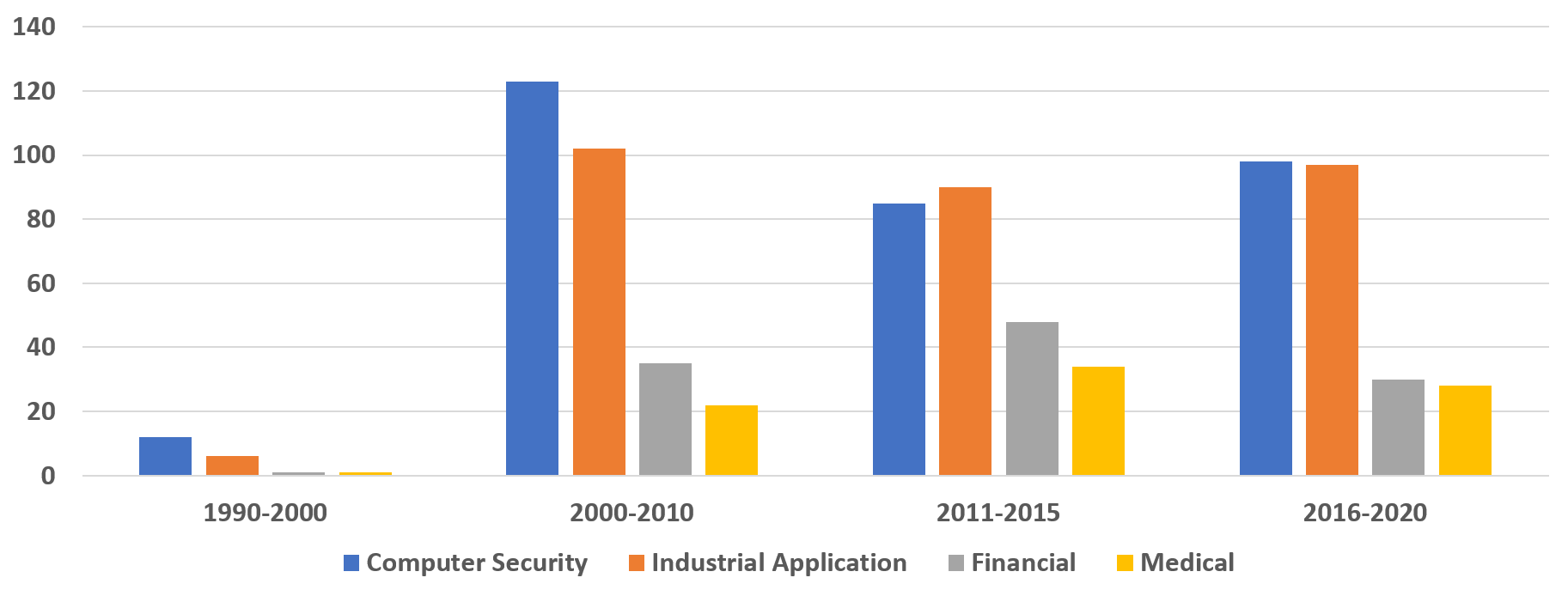}
        \caption{Number of NSA papers found from IEEE Explore and google scholar by year and area of Application keyword search.}
        \label{fig:shorsalgo}
\end{figure}
\begin{figure}
\centering
\subfloat[Pentagon]{
        \includegraphics[width=0.6\linewidth]{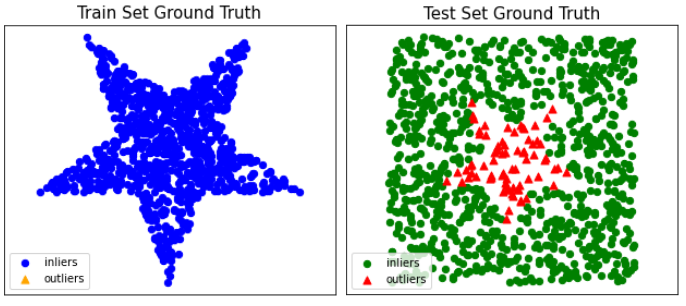}}
        
    \subfloat[Ring]{
        \includegraphics[width=0.6\linewidth]{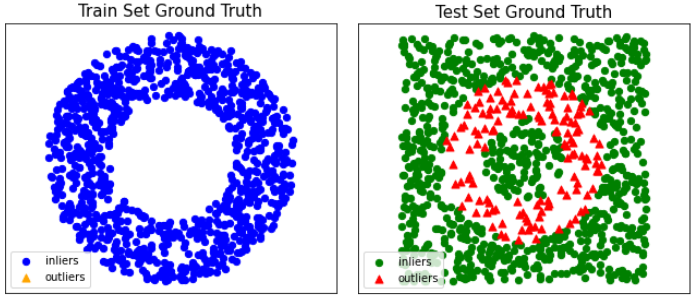} }
   \caption{Ground truth for experimental data}
   \label{fig:trth}
\end{figure}
\begin{figure}
\centering
\subfloat[Pentagon]{
        \includegraphics[width=0.47\linewidth]{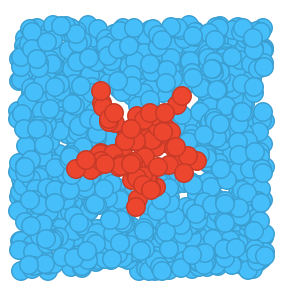}}
    \subfloat[Ring]{
        \includegraphics[width=0.47\linewidth]{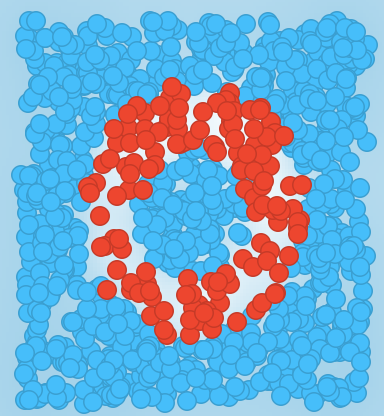} }
   \caption{V-detector (NSA) detection results (blue's are outlier, red are inliers.}
   \label{fig:vdet}
\end{figure}
\begin{table*}
\centering
\tiny
\begin{tabular}{|c|l|l|}
\hline
\textbf{Type}                      & \multicolumn{1}{c|}{\textbf{Abbr}} & \multicolumn{1}{c|}{\textbf{Algorithm}}                                                     \\ \hline
\multirow{3}{*}{Linear Model}      & MCD\cite{hardin2004outlier}                                & Minimum  Covariance Determinant (use the mahalanobis distances as the outlier scores)       \\ \cline{2-3} 
                                   & OCSVM \cite{chen2001one}                             & One-Class   Support Vector Machines                                                         \\ \cline{2-3} 
                                   & LMDD\cite{arning1996linear}                               & Deviation-based  Outlier Detection                                                   \\ \hline
\multirow{6}{*}{Proximity-Based}   & LOF\cite{breunig2000lof}                                & Local Outlier Factor                                                                      \\ \cline{2-3} 
                                   & COF\cite{tang2002enhancing}                                & Connectivity-Based   Outlier Factor                                                         \\ \cline{2-3} 
                                   & CBLOF\cite{he2003discovering}                              & Clustering-Based  Local Outlier Factor                                                     \\ \cline{2-3} 
                                   & HBOS\cite{goldstein2012histogram}                              & Histogram-based   Outlier Score                                                             \\ \cline{2-3} 
                                   & kNN\cite{ramaswamy2000efficient}                                & k   Nearest Neighbors (use the distance to the kth nearest neighbor as the   outlier score) \\ \cline{2-3} 
                                   & SOD \cite{kriegel2009outlier}                               & Subspace   Outlier Detection                                                                \\ \hline
\multirow{3}{*}{Probabilistic}     & ABOD \cite{kriegel2008angle}                              & Angle-Based   Outlier Detection                                                             \\ \cline{2-3} 
                                   & COPOD\cite{li2020copod}                              & COPOD:   Copula-Based Outlier Detection                                                     \\ \cline{2-3}

                                   & SOS\cite{janssens2012stochastic}                               & Stochastic   Outlier Selection                                                              \\ \hline
\multirow{5}{*}{Outlier Ensembles} & IF\cite{togbe2020anomaly}                            & Isolation   Forest                                                                          \\ \cline{2-3} 
                                   & FB \cite{lazarevic2005feature}                                & Feature   Bagging                                                                           \\ \cline{2-3} 
                                   & LSCP \cite{zhao2019lscp}                              & LSCP:   Locally Selective Combination of Parallel Outlier Ensembles                         \\ \cline{2-3} 
                                   & XGBOD \cite{zhao2018xgbod}                             & Extreme   Boosting Based Outlier Detection (Supervised)                                     \\ \cline{2-3} 
                                   & LODA \cite{pevny2016loda}                              & Lightweight   On-line Detector of Anomalies                                                 \\ \hline
\multirow{2}{*}{Neural Networks}   & AutoEncoder\cite{aggarwal2015outlier}                        & Fully   connected AutoEncoder (use reconstruction error as the outlier score)               \\ \cline{2-3} 
                                   & VAE\cite{kingma2013auto}                                & Variational   AutoEncoder (use reconstruction error as the outlier score)                   \\ \hline
\multirow{2}{*}{GAN}               & SO\_GAAL \cite{liu2019generative}                           & Single-Objective   Generative Adversarial Active Learning                                   \\ \cline{2-3} 
                                   & MO\_GAAL\cite{liu2019generative}                             & Multiple-Objective   Generative Adversarial Active Learning                                 \\ \hline
\multirow{3}{*}{NSA}               & Vdetector                          & Variable   Size nagetive selection algorithm                                                \\ \cline{2-3} 
                                   & RNSA                               & Random   real value Negative Selection Algorithm                                            \\ \cline{2-3} 
                                   & GNSA                               & Grid Based   Negative Selection Algorithm                                                   \\ \hline
\end{tabular}
\caption{List of alternate algorithms for NSA (\cite{zhao2019pyod}) and NSA variations we compared in our Experiment.}
\label{tab:alternate}
\end{table*}

NSA method EvoSeedRNSA,  utilizes a Genetic Algorithm (GA) to generate the random seeds to make a sufficient detector set. They consider the detector set as a random sequence made by any random seeds \cite{zhang2014evoseedrnsaii}. EvoSeedRNSA exceeds the established RNSAs and V-detector algorithm. An improved version of this method EvoSeedRNSAII\cite{ZHANG201418} provided more better detector generation using multi-group random seed encoding scheme. 
This method focuses more on the generation of detector initialization but has limitations on the optimization of detector size.
 
 Outlier Robust Negative Selection Algorithm (ORNSA) divided the data into three categories. They are into internal selves, boundary selves, and outlier selves. They use robust outlier detection methods to categorize these three groups, \cite{li2010outlier}. After that, it will join the spaces as a detector space using a positive-negative mechanism. The shortcoming of this method requires self and nonself data to generate detectors. Optimized NSA presents a new NSA adopting an optimization policy stand on re-heating simulated annealing algorithm; this algorithm transforms the position of randomly generated detectors to reach optimal distribution without modifying the number of detectors\cite{aiqiang2011optimization}. This approach has computational limitations while with higher dimensional data-sets.
 An enhanced NSA by integrating a novel different training strategy into the training stage named as FtNSA\cite{farzadnia2020novel}. A new V-detector model named IVRNSA can avoid the time-consuming self-tolerance method of candidate detector inside the coverage of existing older detectors, thus considerably decreasing detector set size and significantly enhancing efficiency\cite{wu2012improved}. CB-RNSA is based on the hierarchical clustering of the self-set, the detection rate of CB-RNSA is higher than that of the classic NSA,  V-detector algorithms, and the false alarm rate is lower than the same algorithms\cite{wen2013negative}. PRR-2NSA is a dual NSA based on pattern recognition receptors theory\cite{zheng2013dual}. The real NSA based on the grid file of feature space (GF-RNSA) aims to improve the exponential worst-case complexity of existing NSA algorithms\cite{wen2014negative}.  
NSA–DE is an improved NSA using differential evolution (DE) optimization with Local outlier factor (LOF), DE is implemented at the random generation phase of NSA, and the LOF  is implemented as a fitness function to maximize the distance of negative and positive space\cite{idris2014hybrid}. Another NSA for Adaptive Network Intrusion Detection System termed as HNSA–IDSA. Here, at the training stage of HNSA–IDSA, both types of data (normal and abnormal) are used to generate normal and abnormal self detectors \cite{ramdane2014new}. NSA–PSO is a novel model that uses particle swarm optimization (PSO) with traditional NSA\cite{idris2015combined}. OALI-detector solves independence difficulty connecting the training stage and testing stage of traditional NSA, and the lack of continuous learning capability makes its detector cannot effectively cover the nonself space\cite{li2015negative}.
Li\cite{li2015negativea} proposed a non-random generation of detectors to detect intrusion. They called their method FB-NSA and FFB-NSA. FB-NSA has two types of detectors: constant-sized detector (CFB-NSA) and variable-sized detector (VFB-NSA) but Due to Non Random generation this technique can be easily by passed.

Immune optimization-based real-valued NSA (IO-RNSA) is based on the self-set distribution in morphological space. IO-RNSA proposes the immune optimization mechanism to generate candidate detectors so that they set from far to near in a hierarchical fashion. With self-sets being the center point, that reduces the repetition of detectors and decreases the numbers of detection holes\cite{xiao2015immune}.   BIORV-NSA\cite{cui2015biorv} overcome some initial NSA defects, such as having too many detector pockets that cannot be identified, and unnecessary invalid detectors are created. OALFB-NSA\cite{li2016boundary} brings the online adaptive learning ability to regular NSA and extends its utilization domains.
Wen\cite{wen2017parameter} proposed a statistical confidence based detector positioning system to improve NSA, the named it DENSA. \cite{pamukov2017multiple} used two negative selection algorithm to detect fault in IoT intrusion detection system, their experiment shows that it reduce false positive rates.Another approach\cite{de2019sensitivity} did a sentiment analysis on relation between detectors neighborhood radius and the number of detectors in order to detect anomalies in data more efficiently. Guerroumi\cite{guerroumi2019nsnad} proposed NSSAD framework which van reduce data-set to improve performance applying a semi-supervised approach and it shows promising results in network anomaly detection. 
\cite{yang2017antigen} used ante gen characteristics to improve NSA, which is improved by Fan\cite{fan2019antigen}, which able to detect coverage using antigen apace triangulation, which performed 10x time better in accuracy than V-detector. Ren\cite{ren2020novel} proposed a new detector overlap calculation model to optimize cost of detector generation. Some notable recent works on NSA include \cite{deng2020negative} which used adaptive immunology in NSA, Selahshoor\cite{selahshoor2019intrusion} used optimize detectors, Song\cite{song2019negative} used in high resistance fault detection, Chikh\cite{chikh2019clustered} used clustered NSA to improve the classification accuracy. 

Some bigdata based approach to make NSA computationally faster is also introduced during this period. Zhu \cite{zhu2017quick} proposed VOR-NSA which can utilize map-reduce efficiently in a distributed system. Hormozi\cite{hormozi2013credit} also uses hadoop to faster the NSA detector size estimation. But the main limitation of VorNSA is use of non-random generation of detectors.

\section{Applications of NSA}

In early age, NSA was mostly limited to computer security and fault detection domain. Recently NSA is mostly used to detect anomaly for different types of data-sets. 

Researchers over the year utilize NSA for solving many problems in diverse kinds of domains. Prominent areas are computer security and fault defections.
\subsection{Computer Security}
Works of NSA in computer security started with 
\cite{forrest1994self} for virus detection and \cite{forrest1996sense} for Unix log anomaly monitor.
NSA uses in Computer security domains are mostly specializes on:
\begin{itemize}
\item Spam Detection: Isris\cite{idris2014improved} combined negative selection algorithm and particle swarm optimization (NSA–PSO) and later\cite{idris2014hybrid} combined NSA and differential evaluations. Saleh\cite{saleh2019intelligent} proposed an elimination based NSA to solve Spam detection problem. Chikh\cite{chikh2019clustered} attempted combination of  fruit fry optimization method with clustered NSA in spam detection.
\item Intrusion Detection: Li\cite{li2015negativea} proposed a non-random generation of detectors to detect intrusion. Another approach\cite{7545821} utilized NSA for distributed network system and shows promising performance.
\item Authentication: Dasgupta\cite{dasgupta2010password} proposed negative filtering for authentication purpose. In their subsequent work\cite{dasgupta2014g} proposed a grid based representation of NSA for authentication. Negative database scheme inspired by NSA used heavily for Authentication purposes \cite{ngo2014largest,hassan2012negative,luo2018authentication}
\item Virus/Malware/Ransomware: Several  researcher\cite{nguyen2014combination,lu2017ransomware,zeng2016extended,wu2014smartphone,zhang2016integrated,brown2016detection,lu2017ransomware} proposed different techniques to detect ransomware, malware or virus. Different data encode method was used along side V-detector provided satisfactory results. 
\end{itemize}

\subsection{Computer Vision}
The researchers relatively neglected the computer vision domain compared to other fields. However, some notable work has been done in this domain with NSA.
Mahapatra\cite{mahapatra2014improved} used the NSA to improve threshold-based segmentation. \cite{ji2006analysis} used NSA to classify dental images. Bendiab\cite{bendiab2010negative} shows that NSA can work for skin image classification.

\subsection{Industrial Engineering} 
Negative selection algorithm is largely used in  fault detection, diagnosis and recovery (FDDR) sector\cite{bayar2015fault}. Most of these works \cite{aydin2010chaotic,laurentys2010design, surace2010novelty, barontini2019deterministically} was done using RRNS and V-detector of NSA.  Alizadeh\cite{alizadeh2017dendritic} Uses sensor fault detection and isolation of wind turbines, similarly \cite{shulin2002negative} uses NSA to detect fault in rotary machines, Abid\cite{abid2017multidomain} uses GA to optimize NSA for bearing fault detection, Guo\cite{guo2019uav} uses a density regulated optimization algorithm with NSA to detect fault in unmanned aerial vehicle sensors, another approach\cite{array2002multilayered} uses multi-layered system for hardware fault detection, \cite{shao2006fault} for fault in gas valve, \cite{ren2006research} for pump-jack fault diagnosis and \cite{xia2007shape} for fault in power transformer. Anomaly detection approach from computer security domain used successfully to detect power grid management\cite{liu2007abnormality} and chemical analysis\cite{cao2003immunogenetic}. \cite{abid2020improved} provided a improved NSA to detect fault but it requires non self data to generate detectors. \cite{jung2017combined} uses residual selection and \cite{song2019negative} proposed an identification framework for fault detection. Another recent work\cite{doi2020early} developed an early warning system for reservoir water release operation using agent-based NSA. The work of Dasgupta\cite{dasgupta2004negative} proved NSA could work to detect the fault in the aircraft control system. Another aircraft based work \cite{chen2020fault} also did similar work, but as they did not state the prior research of \cite{dasgupta2004negative}, their contributions are questionable in terms of novelty.

\subsection{Financial domain}
Ze\cite{ze2008artificial} identify abnormal fluctuation of the stock price using NSA and Butler\cite{butler2010modeling} analyze the behavior of the stock market with NSA.
\cite{hormozi2013credit} added Hadoop processing for NSA in credit card fraud detection. This approach makes the NSA faster and applicable for real-time deployment. Lakshmi\cite{lakshminovel} employed NSA to detect false financial reports.
Mor\cite{mor2011rolling} uses NSA for stock location data analysis and wu\cite{wu2007artificial} implement NSA to identify hedge fund controls over stock market.

\subsection{Medical Domain}
NSA's ability to work without non-self data is a great asset for the medical domain as data bias is a challenging problem in this domain. Many interesting works with NSA have been done in this domain. A novel approach\cite{ba2012eeg} utilized particle swarm optimization with NSA to classify EEG signals. Mousavi\cite{mousavi2013negative} predicted dengue outbreak detection using NSA. Lasisi\cite{lasisi2016application} provided different Application of Real-Valued NSA to improve medical diagnosis. Creevey\cite{creevey2002algorithm} proposed an NSA for protein-coding DNA sequences. Perkins\cite{perkins2019using} applied NSA to identify corrupted Ribo-seq and RNA-seq Samples. Le\cite{lee2004negative} applied NSA for DNA sequence classification. Other works\cite{barontini229negative,li2011hybrid} uses NSA based methodology for online structural health monitoring.


\section{Experiments and Result Analysis with NSA Alternatives}
NSA method has an extensive competition with another alternate approach, which can also work with only self data.
One class classifications (OCC) solve problems where the training datasets only contain samples of one class, and learning models have to identify new data whether belong to that class or not. it is also known as unary classification or class-modelling problem. Most common approach for solving one class problem is one class support vector machine (OCSVM\cite{chen2001one}). Other similar approaches are  Minimum Co-variance Method (MCM) \cite{hubert2018minimum}, Gaussian Mixture Model (GMM)\cite{reynolds2009gaussian}, Dirichlet Process Mixture Model (DPMM)\cite{blei2006variational}, Kernel Density Estimator (KDE)\cite{moon1995estimation}, Robust KDE\cite{kim2012robust}. GWR-Netwrok\cite{marsland2002self}, Deep Support Vector Data Description (SVDD), Deep Auto Encoder based Methods\cite{hinton2006reducing}, Generative Adversarial Net Based approaches (eg: \cite{li2018anomaly,Schlegl2017Unsupervised}, etc. 
These OCC techniques can be classified in 6 types.  Minimum Covariance Determinant(MCD\cite{hardin2004outlier}), OCSVM, Deviation-based(LMDD\cite{arning1996linear}) considered as linear model based OCC techniques. Another type is proximity based which includes
 Local Outlier Factor(LOF)\cite{breunig2000lof},  Connectivity-Based (COF\cite{tang2002enhancing}), Clustering-Based LOF(CBLOF\cite{he2003discovering}),  Histogram-based (HBOS\cite{goldstein2012histogram}), K-Nearest Neighbors (kNN\cite{ramaswamy2000efficient}), Subspace Outlier Detection( SOD \cite{kriegel2009outlier}).        

Angle-Based (ABOD\cite{kriegel2008angle}) , Copula-Based (COPOD\cite{li2020copod}, Stochastic  Selection(SOS\cite{janssens2012stochastic}) are known as probabilistic techniques used for OCC. Combining several methods of OCC are known as ensemble techniques those include Isolation Forest (IF\cite{togbe2020anomaly}), Locally Selective Combination of Parallel Outlier Ensembles(LSCP \cite{zhao2019lscp}), Feature Bagging(FB \cite{lazarevic2005feature}), Extreme Boosting Based (XGBOD \cite{zhao2018xgbod}), etc.

With the  improvement of deep learning methods, OCC problems were solved using different neural network models such as Fully -connected AutoEncoder (AE\cite{aggarwal2015outlier}), Variational  AutoEncoder(VAE\cite{kingma2013auto}), Single-Objective GAN (SO-GAAL \cite{liu2019generative}), Multiple-Objective GAN(MO-GAAL\cite{liu2019generative}), etc.

While Negative selection algorithm (NSA) has several variations, in our experiments we used three variants including variable-size negative selection algorithm(V-detector\cite{ji2005boundary}), Random real-valued Negative Selection Algorithm(RNSA\cite{gonzalez2003randomized}), Grid-Based Negative Selection Algorithm(GNSA\cite{dasgupta2010password}).  

\begin{figure*}
\centering
\subfloat[MCD]{
        \includegraphics[width=0.13\linewidth,height=0.24\linewidth]{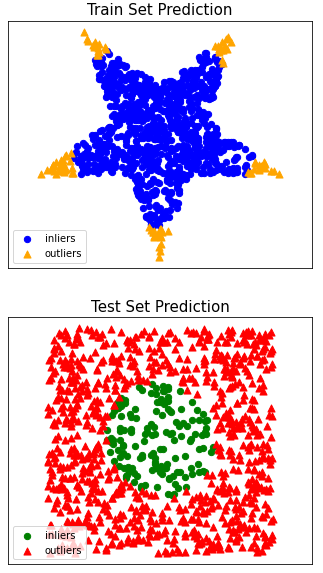}
   }
    \subfloat[OCSVM]{
        \includegraphics[width=0.13\linewidth,height=0.24\linewidth]{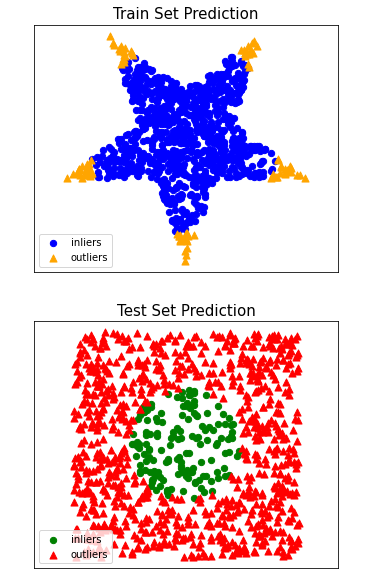}  
    }
    \subfloat[LMDD]{
        \includegraphics[width=0.13\linewidth,height=0.24\linewidth]{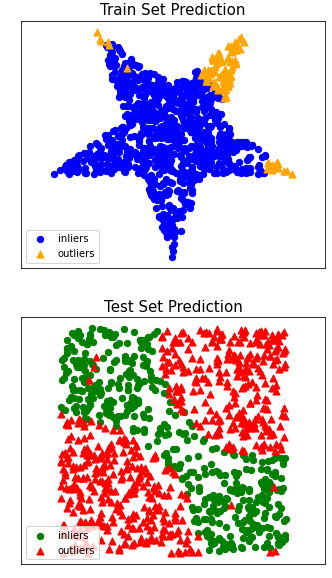}   
   }
    \subfloat[LOF]{
        \includegraphics[width=0.13\linewidth,height=0.24\linewidth]{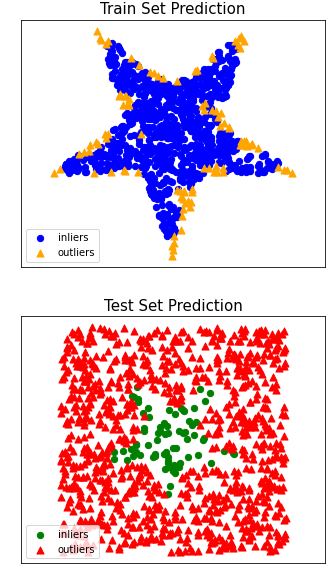}  
    }
    \subfloat[COF]{
        \includegraphics[width=0.13\linewidth,height=0.24\linewidth]{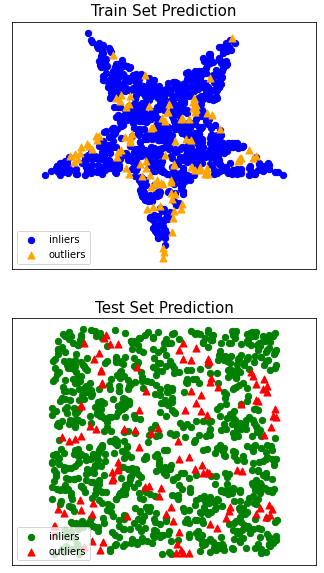}
   }
    \subfloat[COPOD]{
        \includegraphics[width=0.13\linewidth,height=0.24\linewidth]{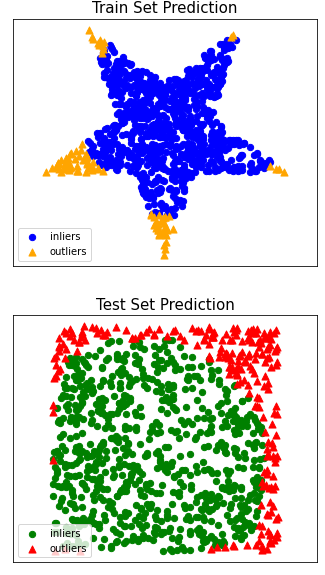}  
    }
    \subfloat[HBOS]{
        \includegraphics[width=0.13\linewidth,height=0.24\linewidth]{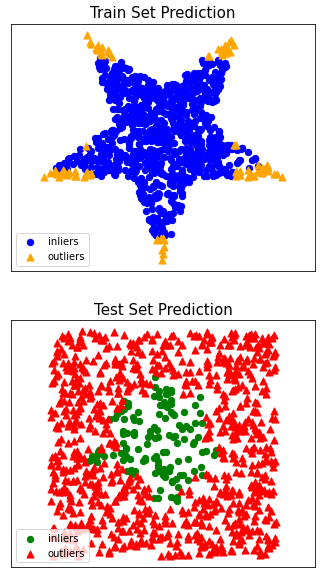}
   }
   
    \subfloat[KNN]{
        \includegraphics[width=0.13\linewidth,height=0.24\linewidth]{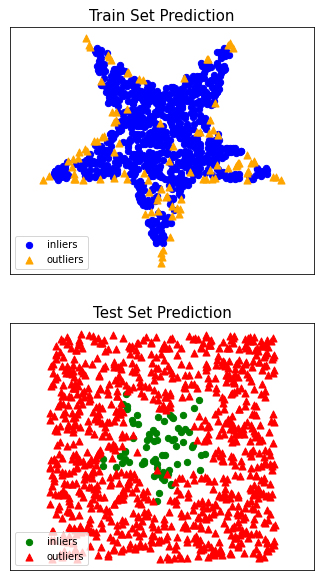}  
    }
       \subfloat[PCA]{
        \includegraphics[width=0.13\linewidth,height=0.24\linewidth]{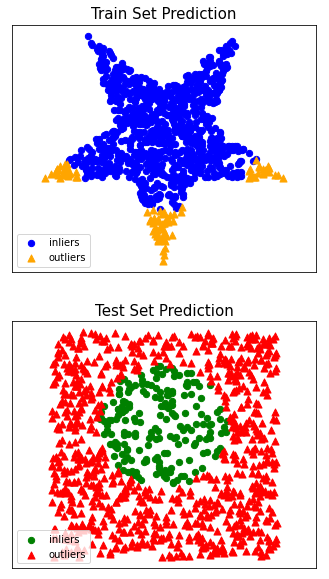}  
    }
      \subfloat[SOD]{
        \includegraphics[width=0.13\linewidth,height=0.24\linewidth]{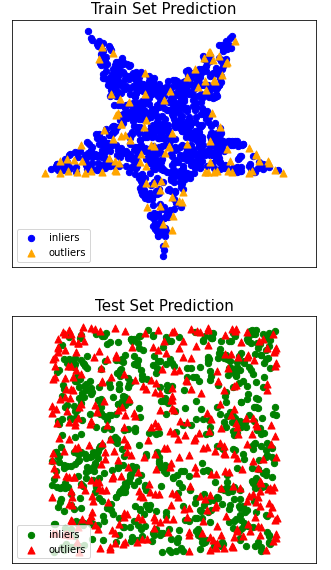}
   }
    \subfloat[ABOD]{
        \includegraphics[width=0.13\linewidth,height=0.24\linewidth]{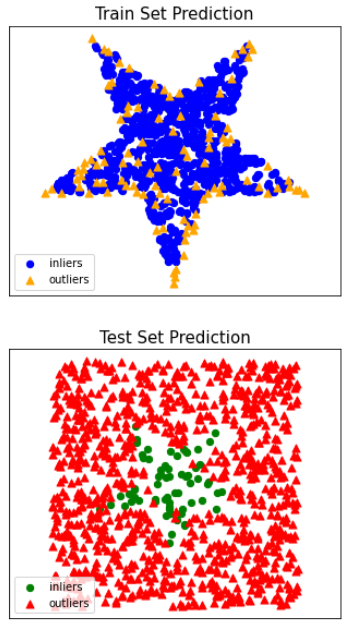}  
    }
    \subfloat[SOS]{
        \includegraphics[width=0.13\linewidth,height=0.24\linewidth]{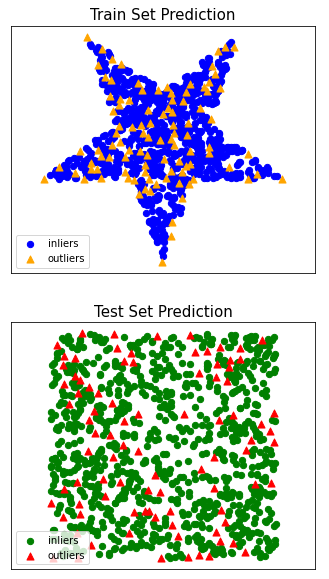}
   }
    \subfloat[IF]{
        \includegraphics[width=0.13\linewidth,height=0.24\linewidth]{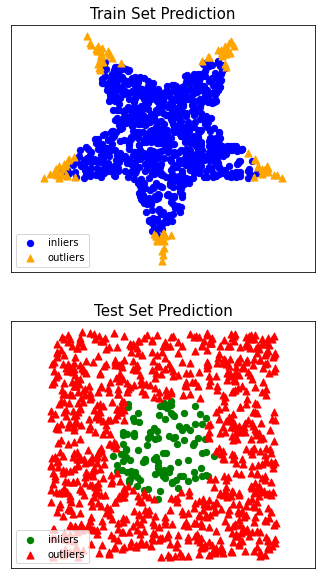}  
    }
    \subfloat[FB]{
        \includegraphics[width=0.13\linewidth,height=0.24\linewidth]{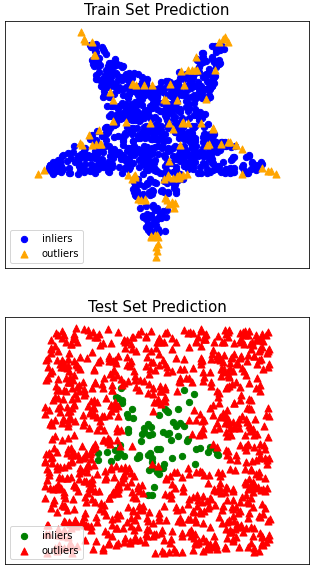}  
        }
        
  \subfloat[LSCP]{
        \includegraphics[width=0.13\linewidth,height=0.24\linewidth]{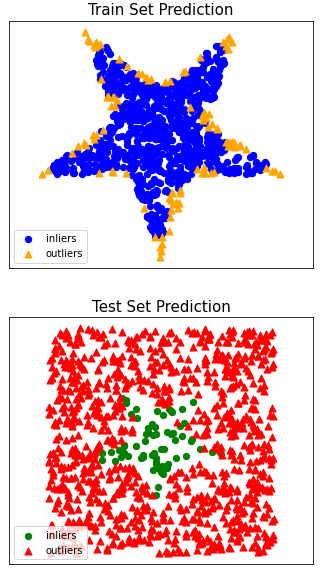}  
        }
        \subfloat[XGBOD]{
        \includegraphics[width=0.13\linewidth,height=0.24\linewidth]{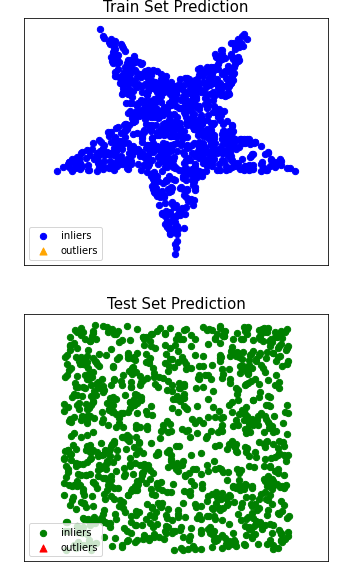} 
        }
         \subfloat[LODA]{
        \includegraphics[width=0.13\linewidth,height=0.24\linewidth]{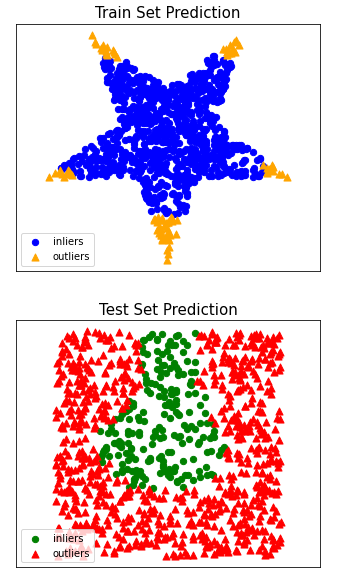} 
        }
           \subfloat[AE]{
        \includegraphics[width=0.13\linewidth,height=0.24\linewidth]{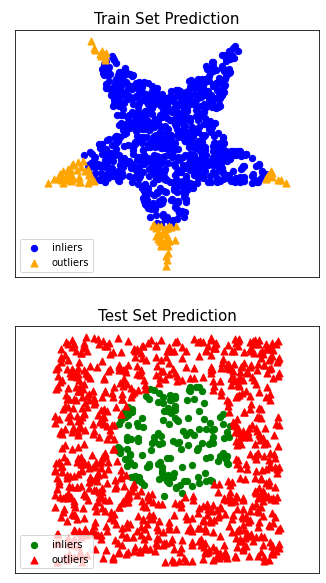} 
        }
           \subfloat[VAE]{
        \includegraphics[width=0.13\linewidth,height=0.24\linewidth]{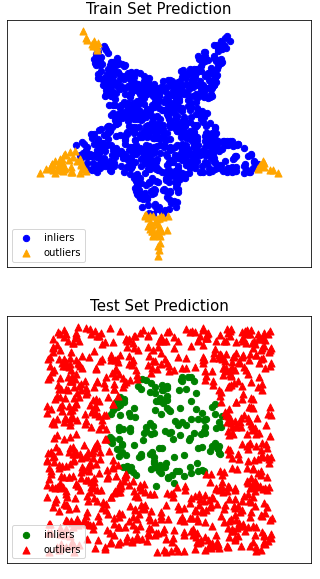}  
        }
         \subfloat[SOGAL]{
        \includegraphics[width=0.13\linewidth,height=0.24\linewidth]{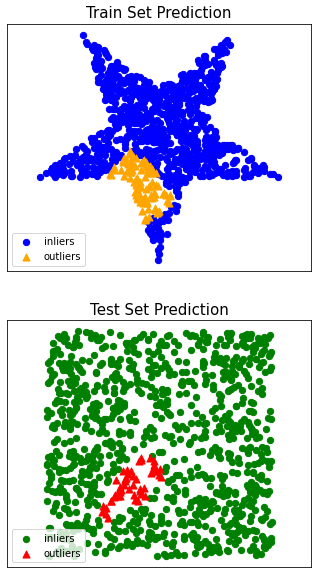} 
        }
                         \subfloat[MOGAL]{
        \includegraphics[width=0.13\linewidth,height=0.24\linewidth]{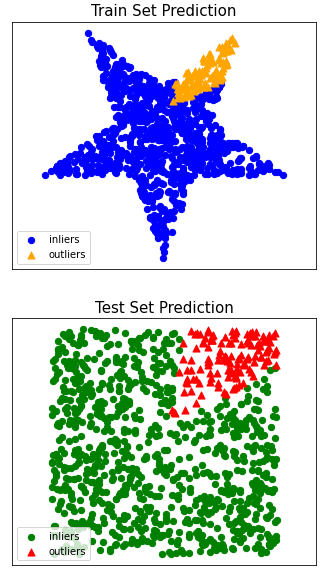} 
        }
        
   \caption{Used Pentagon-shaped  dataset for experimenting with different OCC techniques. Upper image contain the test results on training dataset, blues are correctly identified data and yellow's are mistakenly identified as outlier. In the picture below,  outliers are labelled by red and green are identified as inlier. }
   \label{fig:pentagonresults1}
\end{figure*}

\begin{figure*}
\centering
\subfloat[MCD]{
        \includegraphics[width=0.13\linewidth,height=0.24\linewidth]{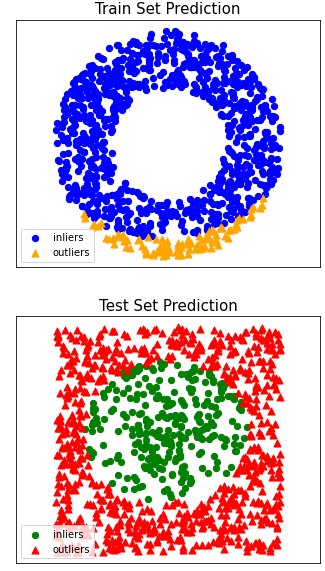}
   }
    \subfloat[OCSVM]{
        \includegraphics[width=0.13\linewidth,height=0.24\linewidth]{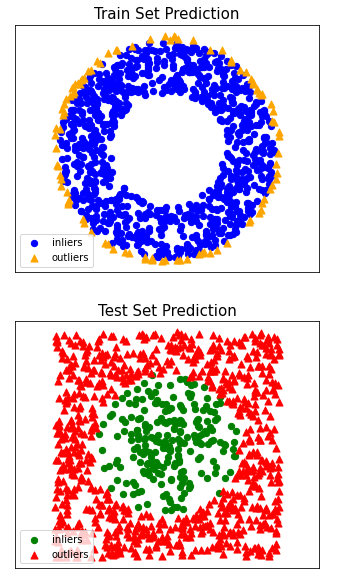}  
    }
    \subfloat[LMDD]{
        \includegraphics[width=0.13\linewidth,height=0.24\linewidth]{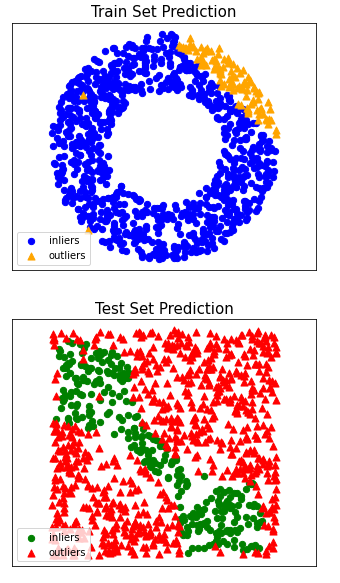}   
   }
    \subfloat[LOF]{
        \includegraphics[width=0.13\linewidth,height=0.24\linewidth]{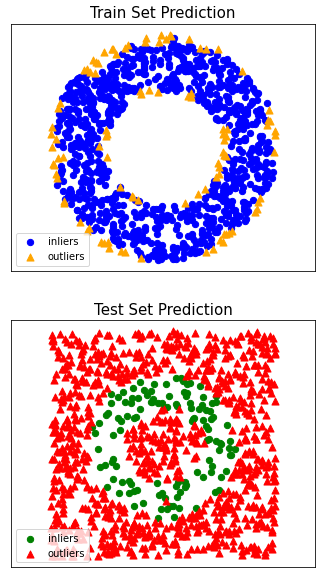}  
    }
    \subfloat[COF]{
        \includegraphics[width=0.13\linewidth,height=0.24\linewidth]{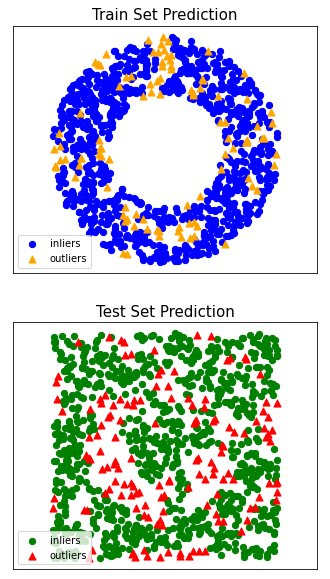}
   }
    \subfloat[COPD]{
        \includegraphics[width=0.13\linewidth,height=0.24\linewidth]{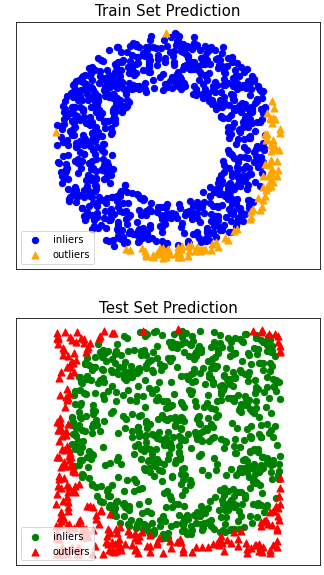}  
    }
    \subfloat[HBOS]{
        \includegraphics[width=0.13\linewidth,height=0.24\linewidth]{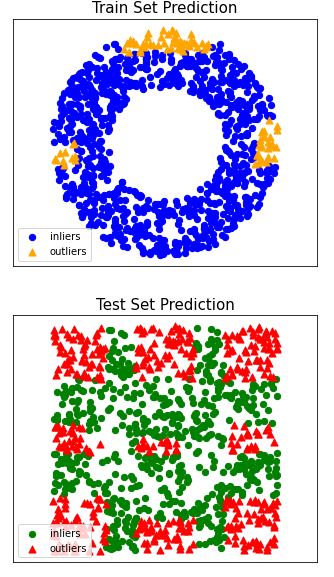}
   }
   
    \subfloat[KNN]{
        \includegraphics[width=0.13\linewidth,height=0.24\linewidth]{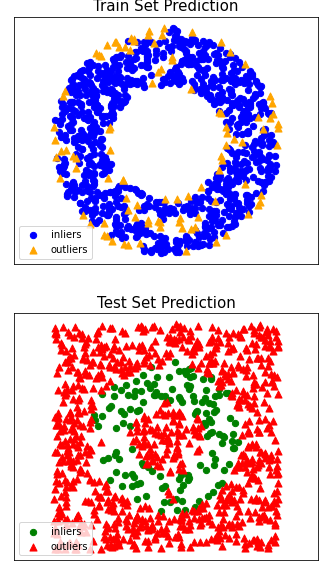}  
    }
       \subfloat[PCA]{
        \includegraphics[width=0.13\linewidth,height=0.24\linewidth]{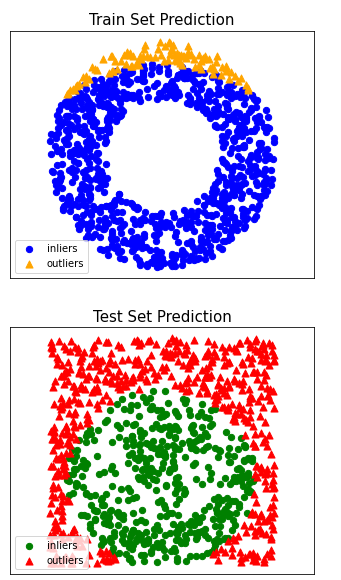}  
    }
      \subfloat[SOD]{
        \includegraphics[width=0.13\linewidth,height=0.24\linewidth]{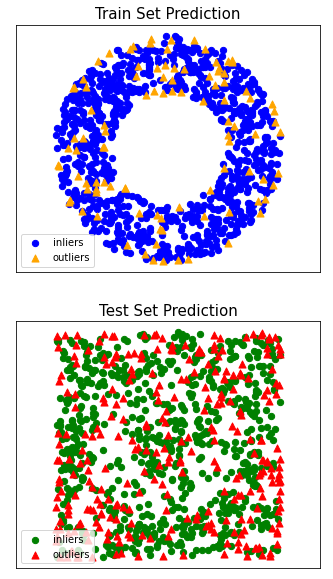}
   }
    \subfloat[ABOD]{
        \includegraphics[width=0.13\linewidth,height=0.24\linewidth]{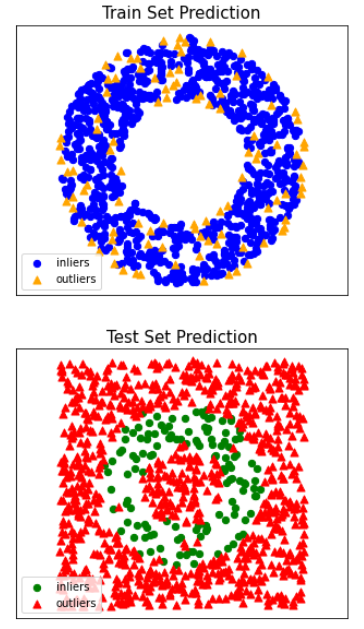}  
    }
    \subfloat[SOS]{
        \includegraphics[width=0.13\linewidth,height=0.24\linewidth]{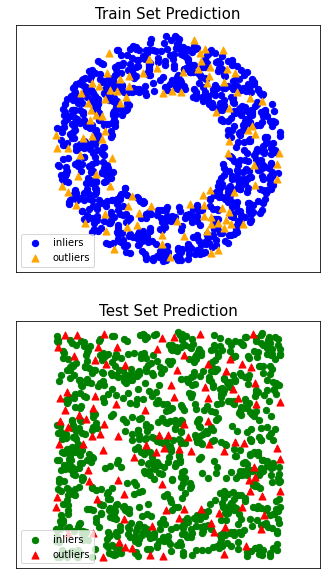}
   }
    \subfloat[IF]{
        \includegraphics[width=0.13\linewidth,height=0.24\linewidth]{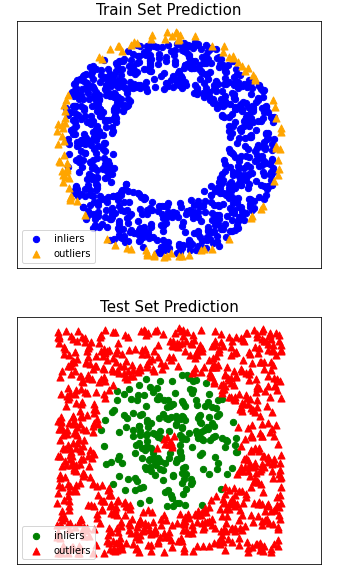}  
    }
    \subfloat[FB]{
        \includegraphics[width=0.13\linewidth,height=0.24\linewidth]{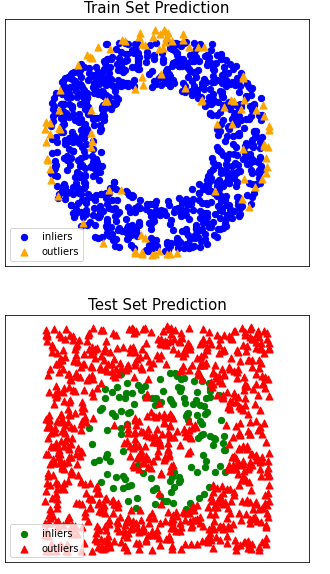}  
        }
        
  \subfloat[LSCP]{
        \includegraphics[width=0.13\linewidth,height=0.24\linewidth]{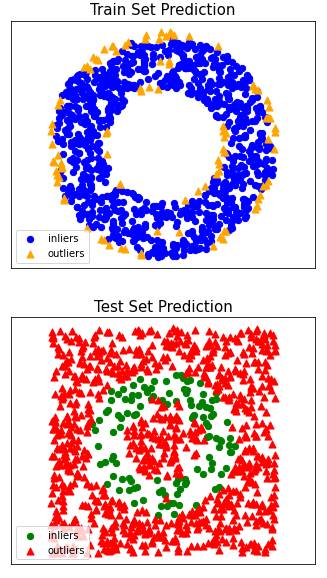}  
        }
        \subfloat[XGBOD]{
        \includegraphics[width=0.13\linewidth,height=0.24\linewidth]{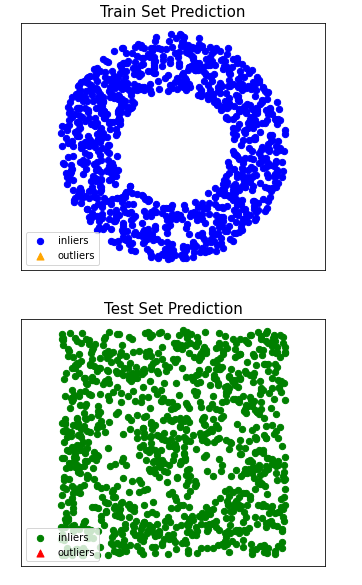} 
        }
         \subfloat[LODA]{
        \includegraphics[width=0.13\linewidth,height=0.24\linewidth]{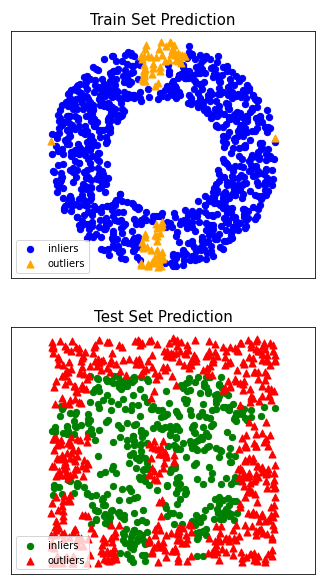} 
        }
           \subfloat[AE]{
        \includegraphics[width=0.13\linewidth,height=0.24\linewidth]{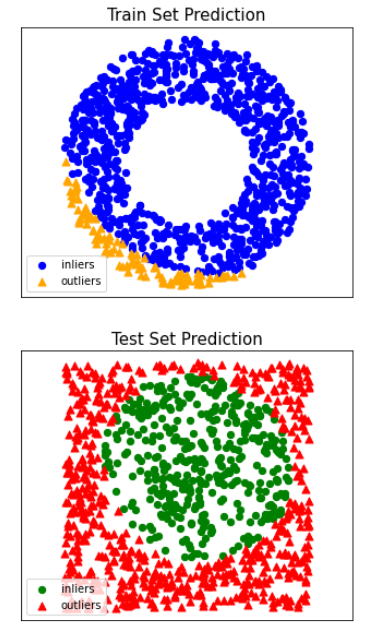} 
        }
           \subfloat[VAE]{
        \includegraphics[width=0.13\linewidth,height=0.24\linewidth]{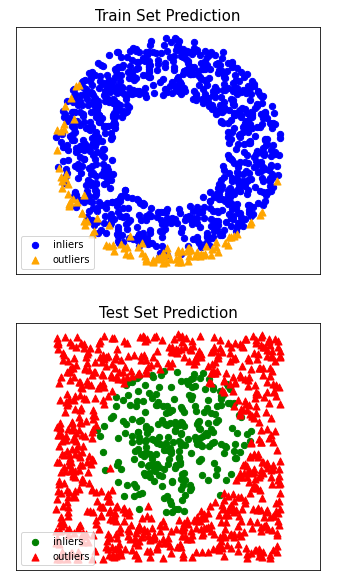}  
        }
         \subfloat[SOGAL]{
        \includegraphics[width=0.13\linewidth,height=0.24\linewidth]{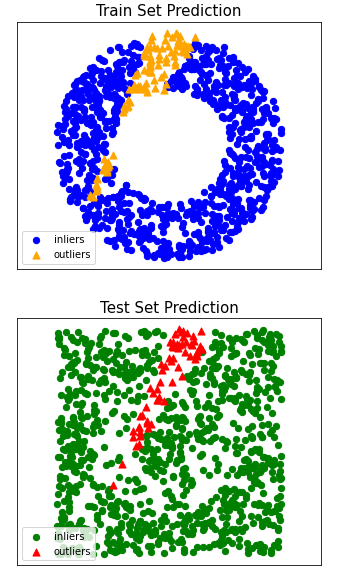} 
        }
                         \subfloat[MOGAL]{
        \includegraphics[width=0.13\linewidth,height=0.24\linewidth]{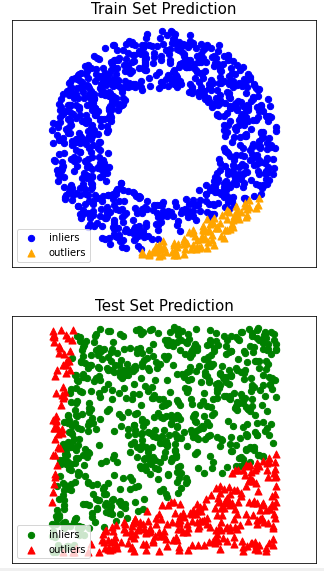} 
        }
        
   \caption{Ring test for different OCC techniques. Upper image contain the test results on train dataset, blues are correctly identified and yellow's are mistakenly identified as outlier, in the below picture red are identified as outlier and green are identified as inlier.}
   \label{fig:pentagonresults}
\end{figure*}

\begin{figure*}
\centering
    \includegraphics[width=1.0\linewidth]{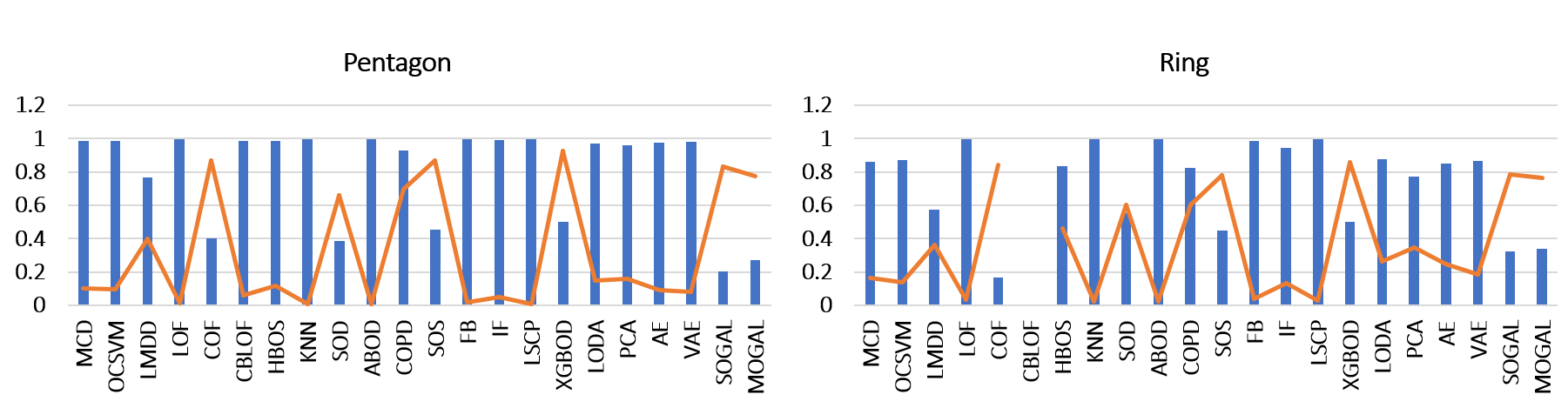}
        \caption{Comparison of different methods error rate (orange line) and ROC (in blue bar)}
        \label{fig:resultcom}
\end{figure*}
\begin{figure}
\centering
    \includegraphics[width=1.0\linewidth]{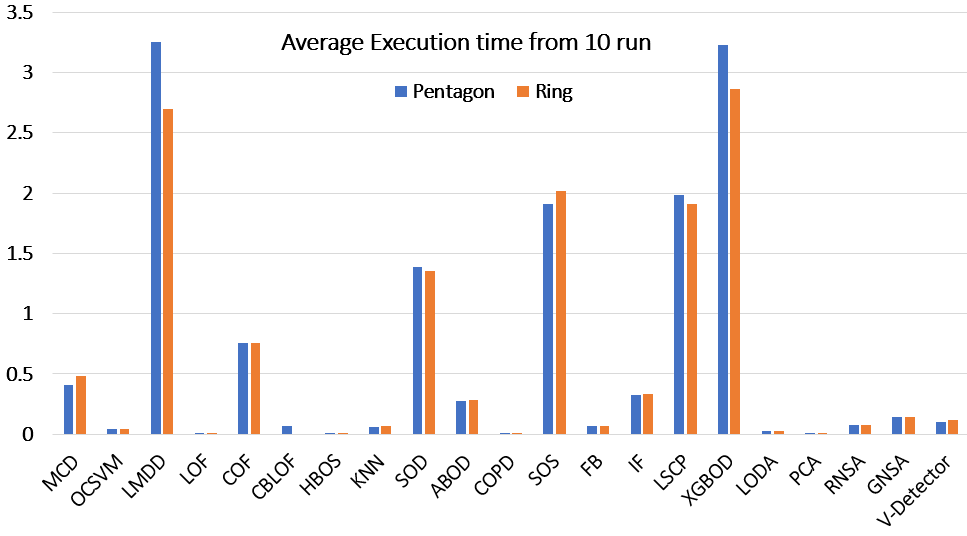}
        \caption{Different methods execution time (neural network and GAN based methods are not presented as they are too high to compare visually)}
        \label{fig:timecom}
\end{figure}

\begin{figure*}
\centering
\subfloat[Arrhythmia]{
       \frame{\includegraphics[width=0.19\linewidth,height=0.24\linewidth]{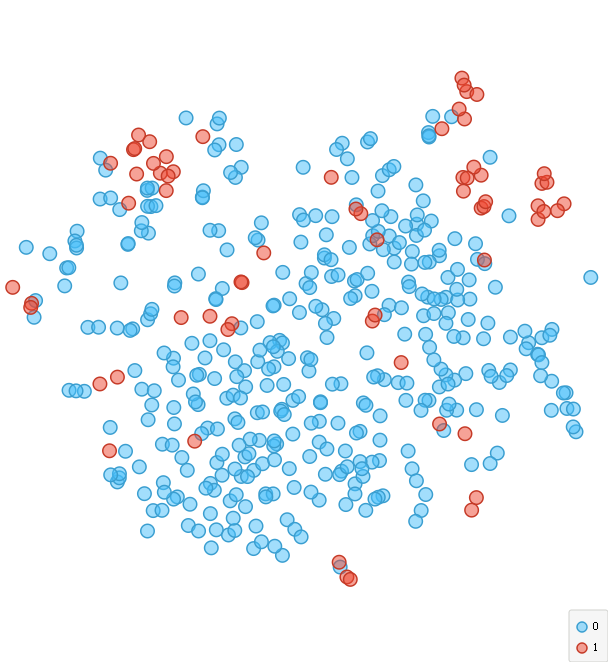}}
   }
    \subfloat[Cardio]{
        \frame{\includegraphics[width=0.19\linewidth,height=0.24\linewidth]{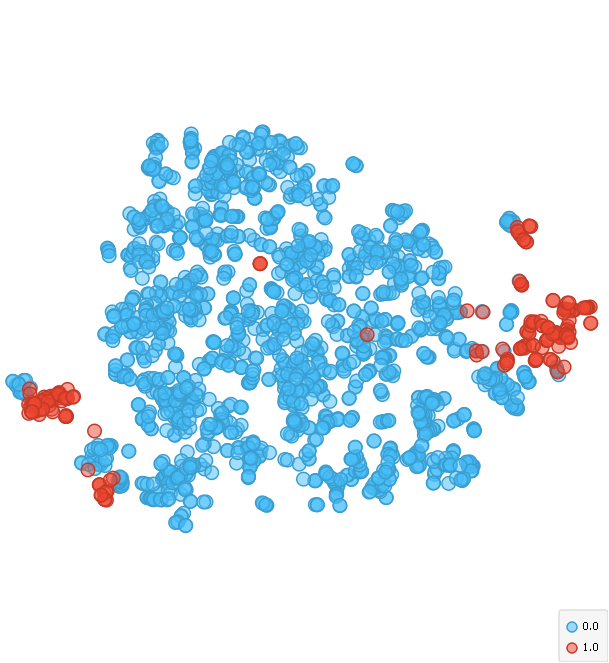}  }
    }
    \subfloat[Glass]{
        \frame{\includegraphics[width=0.19\linewidth,height=0.24\linewidth]{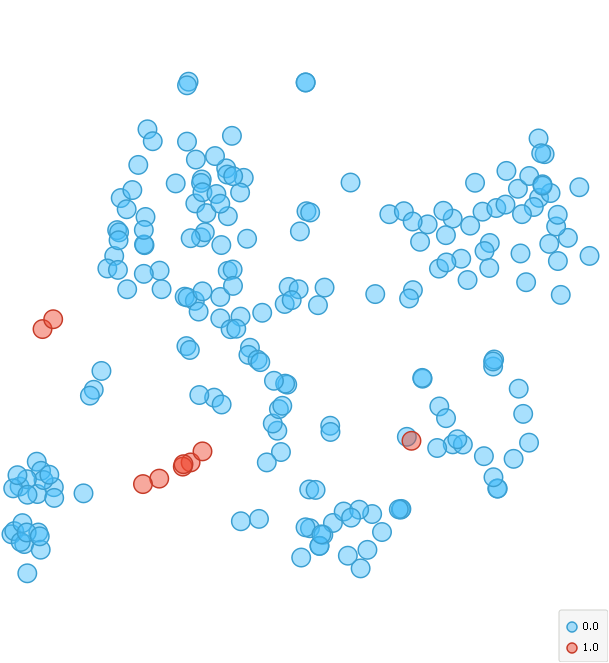}   }
   }
    \subfloat[Ionosphere]{
        \frame{\includegraphics[width=0.19\linewidth,height=0.24\linewidth]{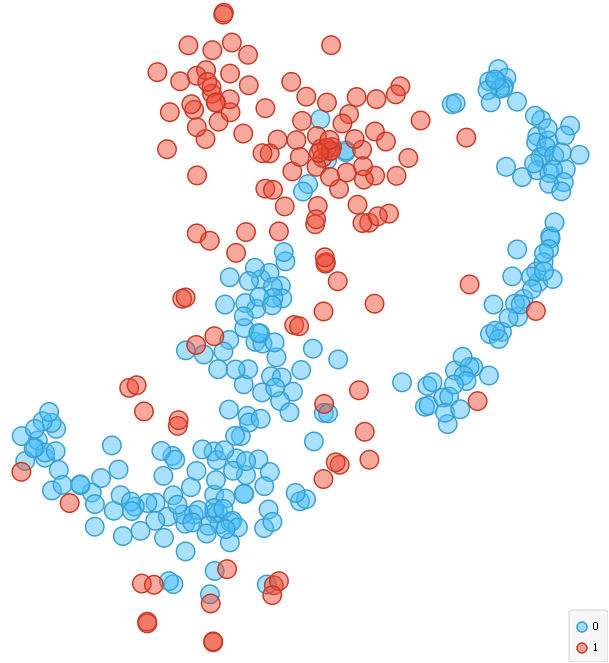}  }
    }
    \subfloat[letter]{
        \frame{\includegraphics[width=0.19\linewidth,height=0.24\linewidth]{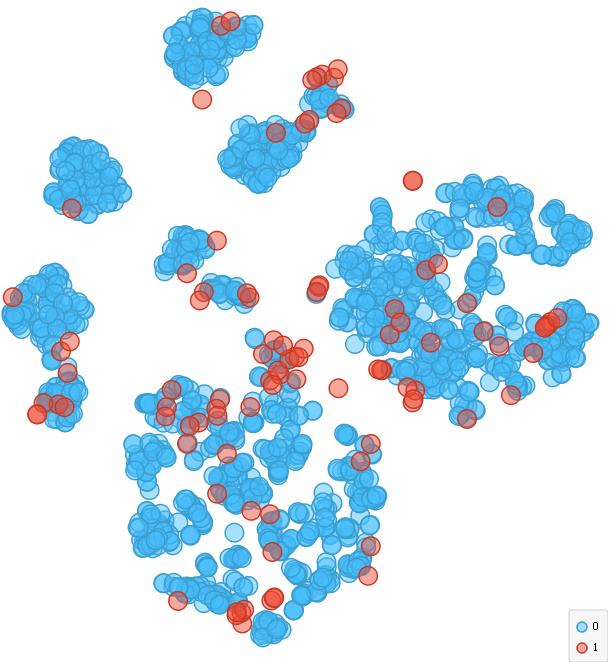}}
   }

    \subfloat[Mnist]{
        \frame{\includegraphics[width=0.19\linewidth,height=0.24\linewidth]{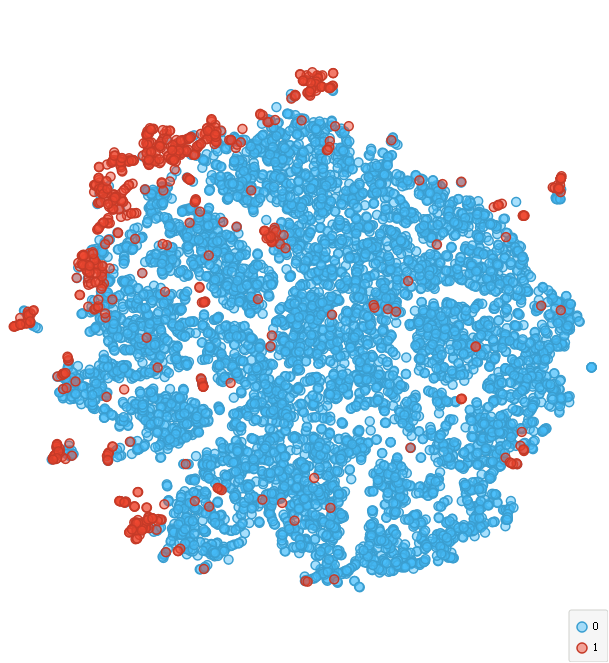}}
   }
    \subfloat[MUSK]{
        \frame{\includegraphics[width=0.19\linewidth,height=0.24\linewidth]{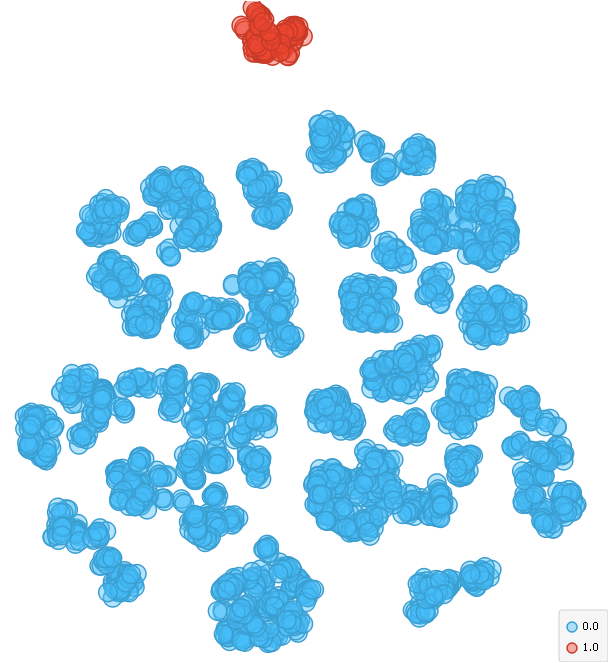}  }
    }
       \subfloat[Optdigits]{
        \frame{\includegraphics[width=0.19\linewidth,height=0.24\linewidth]{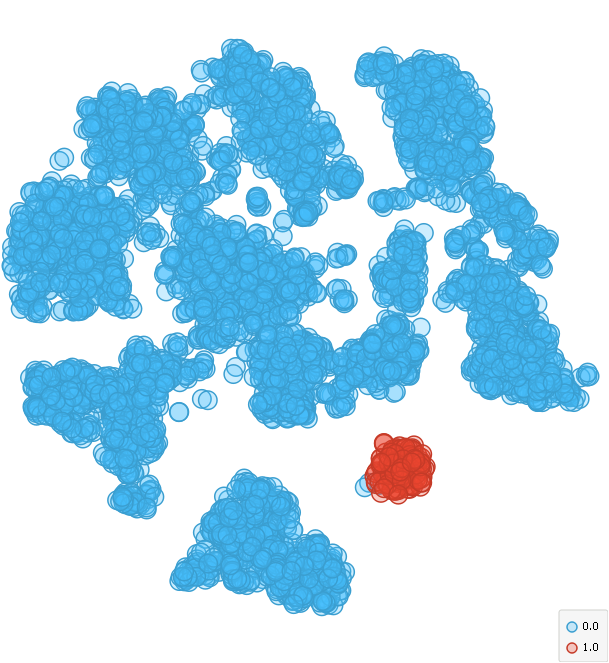}  }
    }
      \subfloat[Pendigits]{
        \frame{\includegraphics[width=0.19\linewidth,height=0.24\linewidth]{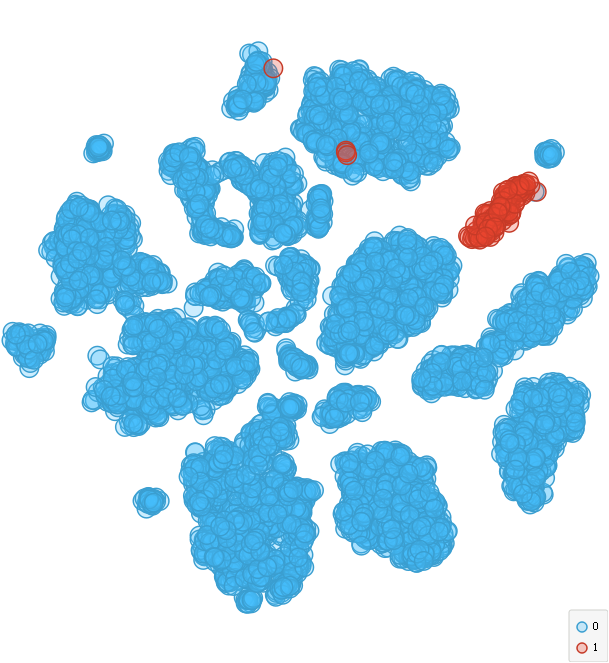}}
   }
    \subfloat[Pima]{
        \frame{\includegraphics[width=0.19\linewidth,height=0.24\linewidth]{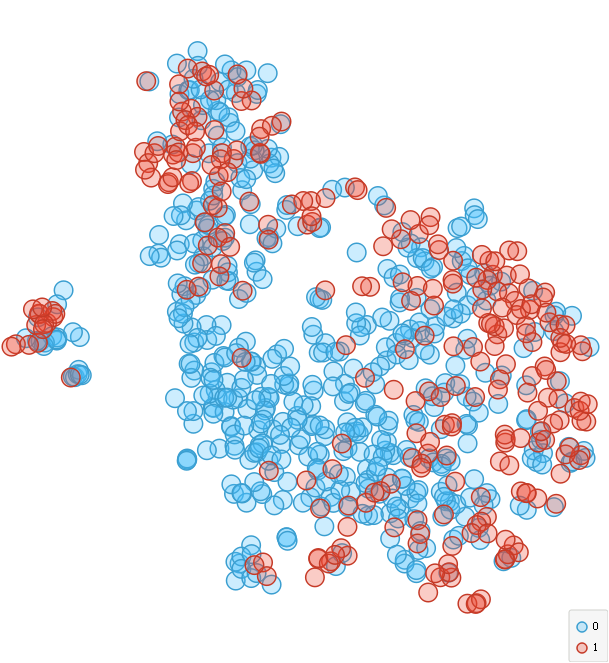}  }
    }
    
    \subfloat[Satelite]{
        \frame{\includegraphics[width=0.19\linewidth,height=0.24\linewidth]{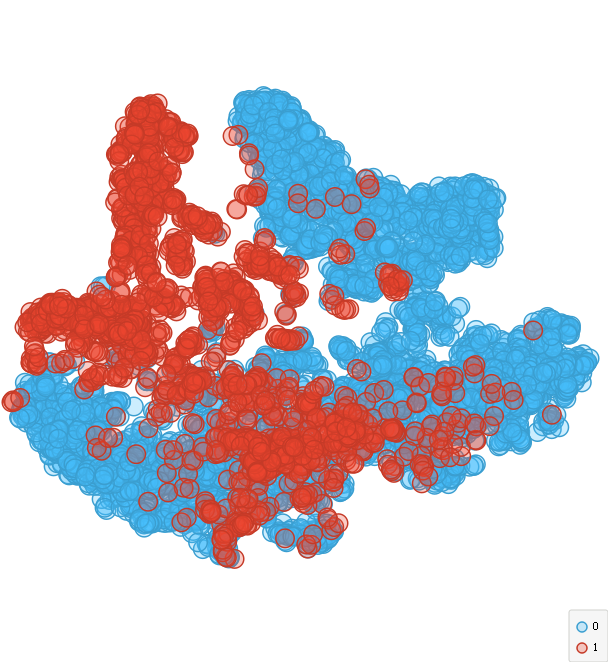}}
   }
    \subfloat[Satimage]{
        \frame{\includegraphics[width=0.19\linewidth,height=0.24\linewidth]{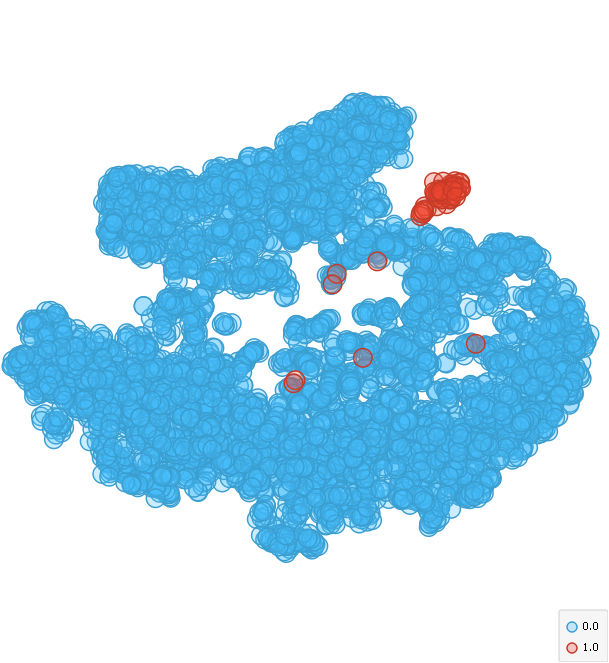} } 
    }
    \subfloat[Shuttle]{
        \frame{\includegraphics[width=0.19\linewidth,height=0.24\linewidth]{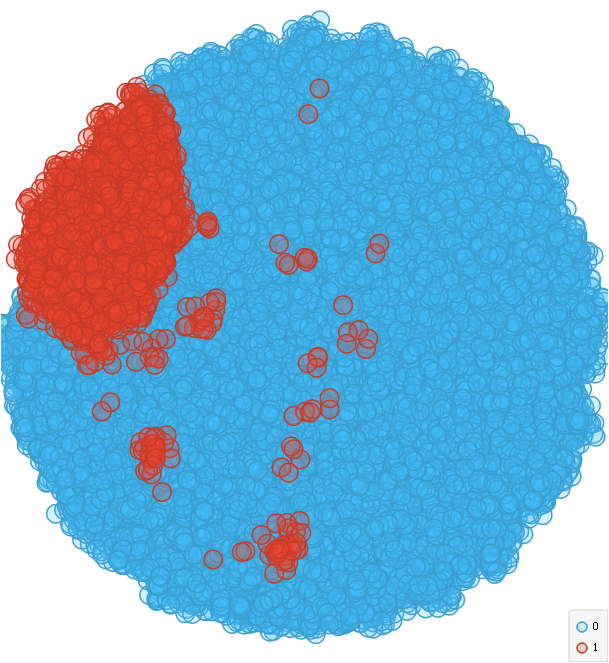}  }
        }
  \subfloat[Veterni]{
        \frame{\includegraphics[width=0.19\linewidth,height=0.24\linewidth]{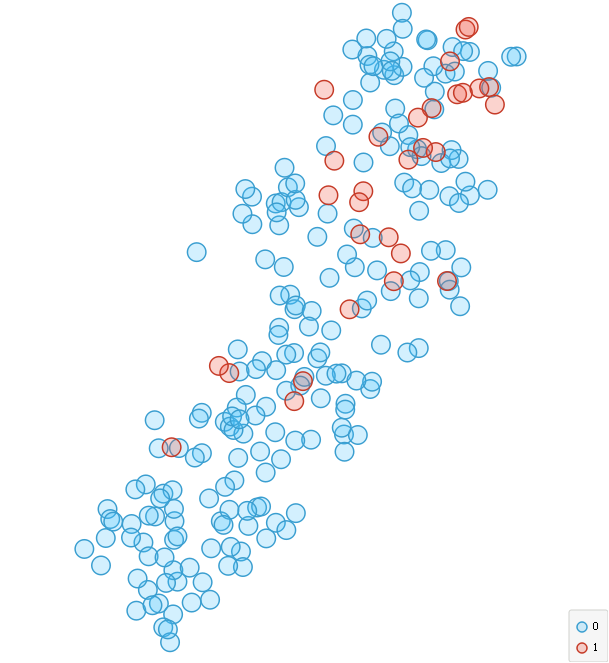}  }
        }
        \subfloat[Vowels]{
        \frame{\includegraphics[width=0.19\linewidth,height=0.24\linewidth]{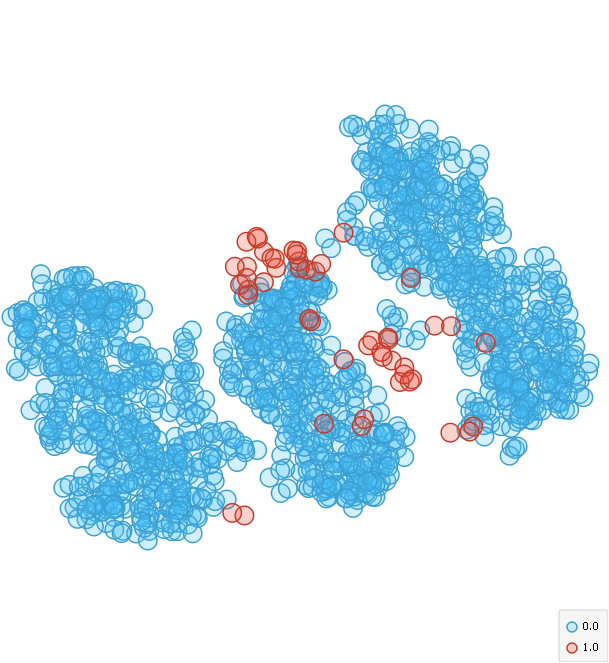} }
        }
        
   \caption{ One class classification datasets representation in 2d using t-SNE representation (blue's are inliers , red are outliers).}
   \label{fig:comparisonfigse}
\end{figure*}
\begin{table}[ht]
\centering
\begin{tabular}{|l|l|l|l|l|l|l|}
\hline
\multicolumn{1}{|c|}{\multirow{2}{*}{{ Model}}} & \multicolumn{3}{c|}{{ \textbf{Pentagon}}} & \multicolumn{3}{c|}{{ \textbf{Ring}}} \\ \cline{2-7} 
\multicolumn{1}{|c|}{}                             & \textbf{ROC}   & \textbf{ER}  & \textbf{ET}  & \textbf{ROC} & \textbf{ER} & \textbf{ET} \\ \hline
\textbf{MCD}                                       & 0.9846         & 0.099        & 0.4101       & 0.8616       & 0.164       & 0.4871      \\ \hline
\textbf{OCSVM}                                     & 0.9851         & 0.097        & 0.0421       & 0.8731       & 0.135       & 0.0417      \\ \hline
\textbf{LMDD}                                      & 0.7673         & 0.401        & 3.253        & 0.5752       & 0.361       & 2.6965      \\ \hline
\textbf{LOF}                                       & 0.997          & 0.012        & 0.0083       & 0.9963       & 0.032       & 0.0096      \\ \hline
\textbf{COF}                                       & 0.3991         & 0.87         & 0.7568       & 0.168        & 0.841       & 0.7602      \\ \hline
\textbf{CBLOF}                                     & 0.9866         & 0.059        & 0.0698       &              &             &             \\ \hline
\textbf{HBOS}                                      & 0.9865         & 0.115        & 0.0015       & 0.8354       & 0.459       & 0.0016      \\ \hline
\textbf{KNN}                                       & 0.9993         & 0.009        & 0.0628       & 0.9957       & 0.025       & 0.0682      \\ \hline
\textbf{SOD}                                       & 0.3842         & 0.661        & 1.3831       & 0.554        & 0.603       & 1.3571      \\ \hline
\textbf{ABOD}                                      & 0.9994         & 0.009        & 0.2776       & 0.9982       & 0.022       & 0.2881      \\ \hline
\textbf{COPD}                                      & 0.9273         & 0.696        & 0.0105       & 0.8255       & 0.603       & 0.0099      \\ \hline
\textbf{SOS}                                       & 0.4551         & 0.87         & 1.9094       & 0.4504       & 0.779       & 2.0156      \\ \hline
\textbf{FB}                                        & 0.9942         & 0.019        & 0.0692       & 0.9863       & 0.039       & 0.0716      \\ \hline
\textbf{IF}                                        & 0.9933         & 0.047        & 0.327        & 0.9444       & 0.134       & 0.3339      \\ \hline
\textbf{LSCP}                                      & 0.9992         & 0.009        & 1.9832       & 0.9982       & 0.027       & 1.9078      \\ \hline
\textbf{XGBOD}                                     & 0.5            & 0.926        & 3.2267       & 0.5          & 0.859       & 2.8597      \\ \hline
\textbf{LODA}                                      & 0.9703         & 0.149        & 0.029        & 0.8766       & 0.261       & 0.0286      \\ \hline
\textbf{PCA}                                       & 0.9588         & 0.158        & 0.0017       & 0.7747       & 0.348       & 0.0019      \\ \hline
\textbf{AE}                                        & 0.9738         & 0.093        & 4.62         & 0.8511       & 0.247       & 4.4478      \\ \hline
\textbf{VAE}                                       & 0.9833         & 0.08         & 5.5          & 0.8642       & 0.182       & 5.1877      \\ \hline
\textbf{SOGAL}                                     & 0.2002         & 0.831        & 11.6         & 0.3248       & 0.785       & 10.683      \\ \hline
\textbf{MOGAL}                                     & 0.2716         & 0.774        & 84           & 0.3361       & 0.762       & 95.4466     \\ \hline
\textbf{RNSA}                                       & 0.92         & 0.7	         &0.08          & 0.97       & 0.2       & 0.08     \\ \hline
\textbf{VDetecor}                                     & 0.98        & 0.09        & 0.14        & 0.99       & 0       & 0.14     \\ \hline
\textbf{GNSA}                                     & 0.76         & 0.43        & 0.1          & 0.6      & 0.52       & 0.13   \\ \hline
\end{tabular}
\caption{Result Comparison of different models(ER: Error Rate, ET: Execution time)}
\label{tab:results}
\end{table}
\begin{table*}[ht]
\centering
\begin{tabular}{|l|l|l|l|l|l|l|l|l|l|l|l|}
\hline
\multicolumn{1}{|c|}{data} & \multicolumn{1}{c|}{\textbf{ABOD}} & \multicolumn{1}{c|}{\textbf{CBLOF}} & \multicolumn{1}{c|}{\textbf{FB}} & \multicolumn{1}{c|}{\textbf{HBOS}} & \multicolumn{1}{c|}{\textbf{IForest}} & \multicolumn{1}{c|}{\textbf{KNN}} & \multicolumn{1}{c|}{\textbf{LOF}} & \multicolumn{1}{c|}{\textbf{MCD}} & \multicolumn{1}{c|}{\textbf{OCSVM}} & \multicolumn{1}{c|}{\textbf{PCA}} & \multicolumn{1}{c|}{\textbf{V-NSA}} \\ \hline
\textbf{Arrhythmia}        & 0.7688                             & 0.7835                              & 0.7781                           & 0.8219                             & 0.8005                                & 0.7861                            & 0.7787                            & 0.779                             & 0.7812                              & 0.7815                            & 0.7                                      \\ \hline
\textbf{Cardio}            & 0.5692                             & 0.9276                              & 0.5867                           & 0.8351                             & 0.9213                                & 0.7236                            & 0.5736                            & 0.8135                            & 0.9348                              & 0.9504                            & 0.9                                      \\ \hline
\textbf{Glass}             & 0.7951                             & 0.8504                              & 0.8726                           & 0.7389                             & 0.7569                                & 0.8508                            & 0.8644                            & 0.7901                            & 0.6324                              & 0.6747                            & 0.9                                      \\ \hline
\textbf{Ionosphere}        & 0.9248                             & 0.8134                              & 0.873                            & 0.5614                             & 0.8499                                & 0.9267                            & 0.8753                            & 0.9557                            & 0.8419                              & 0.7962                            & 0.9                                      \\ \hline
\textbf{Letter}            & 0.8783                             & 0.507                               & 0.866                            & 0.5927                             & 0.642                                 & 0.8766                            & 0.8594                            & 0.8074                            & 0.6118                              & 0.5283                            & 0.7                                      \\ \hline
\textbf{Lympho}            & 0.911                              & 0.9728                              & 0.9753                           & 0.9957                             & 0.9941                                & 0.9745                            & 0.9771                            & 0.9                               & 0.9759                              & 0.9847                            & 1                                        \\ \hline
\textbf{Mnist}             & 0.7815                             & 0.8009                              & 0.7205                           & 0.5742                             & 0.8159                                & 0.8481                            & 0.7161                            & 0.8666                            & 0.8529                              & 0.8527                            & 0.9                                      \\ \hline
\textbf{Musk}              & 0.1844                             & 0.9879                              & 0.5263                           & 1                                  & 0.9999                                & 0.7986                            & 0.5287                            & 0.9998                            & 1                                   & 1                                 & 1                                        \\ \hline
\textbf{Optdigits}         & 0.4667                             & 0.5089                              & 0.4434                           & 0.8732                             & 0.7253                                & 0.3708                            & 0.45                              & 0.3979                            & 0.4997                              & 0.5086                            & 0.6                                      \\ \hline
\textbf{Pendigits}         & 0.6878                             & 0.9486                              & 0.4595                           & 0.9238                             & 0.9435                                & 0.7486                            & 0.4698                            & 0.8344                            & 0.9303                              & 0.9352                            & 0.99                                      \\ \hline
\textbf{Pima}              & 0.6794                             & 0.7348                              & 0.6235                           & 0.7                                & 0.6806                                & 0.7078                            & 0.6271                            & 0.6753                            & 0.6215                              & 0.6481                            & 0.4                                      \\ \hline
\textbf{Satellite}         & 0.5714                             & 0.6693                              & 0.5572                           & 0.7581                             & 0.7022                                & 0.6836                            & 0.5573                            & 0.803                             & 0.6622                              & 0.5988                            & 0.7                                      \\ \hline
\textbf{Satimage-2}        & 0.819                              & 0.9917                              & 0.457                            & 0.9804                             & 0.9947                                & 0.9536                            & 0.4577                            & 0.9959                            & 0.9978                              & 0.9822                            & 0.94                                        \\ \hline
\textbf{Shuttle}           & 0.6234                             & 0.6272                              & 0.4724                           & 0.9855                             & 0.9971                                & 0.6537                            & 0.5264                            & 0.9903                            & 0.9917                              & 0.9898                            & 0.94                                        \\ \hline
\textbf{Vertebral}         & 0.4262                             & 0.3486                              & 0.4166                           & 0.3263                             & 0.3905                                & 0.3817                            & 0.4081                            & 0.3906                            & 0.4431                              & 0.4027                            & 0.5                                      \\ \hline
\textbf{Vowels}            & 0.9606                             & 0.5856                              & 0.9425                           & 0.6727                             & 0.7585                                & 0.968                             & 0.941                             & 0.8076                            & 0.7802                              & 0.6027                            & 0.8                                      \\ \hline
\textbf{Wbc}               & 0.9047                             & 0.9227                              & 0.9325                           & 0.9516                             & 0.931                                 & 0.9366                            & 0.9349                            & 0.921                             & 0.9319                              & 0.9159                            & 1                                        \\ \hline
\textbf{Average}               & 0.7030                           & 0.7635                            & 0.6766                         & 0.7818                           & 0.8178                              & 0.7758                          & 0.6791                          & 0.8075                          & 0.7934                            & 0.7736                          & 0.77                                 \\ \hline
\end{tabular}
\caption{NSA results compared with other OCC methods for different datasets. Here NSA experiments were conducted after  t-distributed stochastic neighbourhood embedding (t-SNE \cite{van2008visualizing}) dimension reduction and the results of other methods were reported from \cite{zhao2019pyod} experiments.}
\label{tab:otherdatasets}
\end{table*}

\begin{figure}
\centering
    \includegraphics[width=1.0\linewidth]{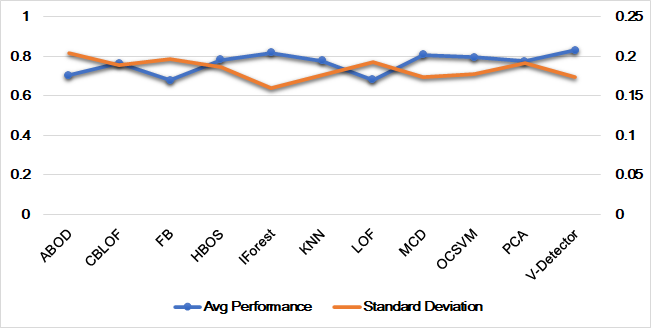}
        \caption{Results of different OCC methods (average and standard deviation) for all 17 datasets in terms of performance accuracy.}
        \label{fig:rr}
\end{figure}

\subsection{One class classifications (OCC)}
One class classifications (OCC) solve problems where the training datasets only contain samples of one class, and learning models have to identify new data whether belong to that class or not. it is also known as unary classification or class-modelling problem. Most common approach for solving one class problem is one class support vector machine (OCSVM\cite{chen2001one}). Other similar approaches are  Minimum Co-variance Method (MCM) \cite{hubert2018minimum}, Gaussian Mixture Model (GMM)\cite{reynolds2009gaussian}, Dirichlet Process Mixture Model (DPMM)\cite{blei2006variational}, Kernel Density Estimator (KDE)\cite{moon1995estimation}, Robust KDE\cite{kim2012robust}. GWR-Netwrok\cite{marsland2002self}, Deep Support Vector Data Description (SVDD), Deep Auto Encoder based Methods\cite{hinton2006reducing}, Generative Adversarial Net Based approaches (eg: \cite{li2018anomaly,Schlegl2017Unsupervised}, etc. 
These OCC techniques can be classified in 6 types.  Minimum Covariance Determinant(MCD\cite{hardin2004outlier}), OCSVM, Deviation-based(LMDD\cite{arning1996linear}) considered as linear model based OCC techniques. Another type is proximity based which includes
 Local Outlier Factor(LOF)\cite{breunig2000lof},  Connectivity-Based (COF\cite{tang2002enhancing}), Clustering-Based LOF(CBLOF\cite{he2003discovering}),  Histogram-based (HBOS\cite{goldstein2012histogram}), K-Nearest Neighbors (kNN\cite{ramaswamy2000efficient}), Subspace Outlier Detection( SOD \cite{kriegel2009outlier}).        

Angle-Based (ABOD\cite{kriegel2008angle}) , Copula-Based (COPOD\cite{li2020copod}, Stochastic  Selection(SOS\cite{janssens2012stochastic}) are known as probabilistic techniques used for OCC. Combining several methods of OCC are known as ensemble techniques those include Isolation Forest (IF\cite{togbe2020anomaly}), Locally Selective Combination of Parallel Outlier Ensembles(LSCP \cite{zhao2019lscp}), Feature Bagging(FB \cite{lazarevic2005feature}), Extreme Boosting Based (XGBOD \cite{zhao2018xgbod}), etc.

With the  improvement of deeplearning methods, OCC problems were solved using different neural network models such as Fully connected AutoEncoder (AE\cite{aggarwal2015outlier}), Variational  AutoEncoder(VAE\cite{kingma2013auto}), Single-Objective GAN (SO-GAAL \cite{liu2019generative}), Multiple-Objective GAN(MO-GAAL\cite{liu2019generative}), etc.

While Negative selection algorithm (NSA) has several variations, in our experiments we used three variants including variable-size negative selection algorithm(V-detector\cite{ji2005boundary}), Random real-valued Negative Selection Algorithm(RNSA\cite{gonzalez2003randomized}), Grid-Based Negative Selection Algorithm(GNSA\cite{dasgupta2010password}).

\subsection{Experiment Methodology}
We focused to visualize the impact of different OCC methods with NSA variations, and to best visualize, we took shape data-set. Inside the shapes is the true class, and outside the shape would be the false class. We have 2000 data points inside the shape class. We have another data-set containing 2000 data-point inside and outside shapes for the test.  We experimented with two different shapes; Ring and Pentagon. There is inside negative and outside the positive region in the ring, and for the pentagon, outside detection is not linearly separable. In figure \ref{fig:trth} illustrated the inlier and outlier position for Ring and Pentagon test and train data. This nature of the shape made it possible to experiment and better visualize. For conducting experiments, we used python pyod library \cite{zhao2019pyod}, and for NSA variations, we used Zhou-Ji's developed code\cite{Zhou-06}. Also used the default configuration of pyod library, and for NSA methods, considered minkowski distance. Our experimental platform was google co-lab with GPU and high ram usage enabled. Also, we used open-source repository of data and source code for reproducibility. For CBLOF, the Ring dataset was not considered. We used pyod built in visualization utility and orange tool\cite{demvsar2013orange} for visual image generation.

For more detailed evaluation, we experimented with 17 Outlier Detection DataSets (ODDS) from different domains (collected from odds lab, stonybrook university\urlfootnote{http://odds.cs.stonybrook.edu}). These datasets have one record per data point, and each record contains several attributes. As these datasets have a low number of samples, they are not suitable for neural network and GAN-based approaches. We presented other ten different OCC techniques from \cite{zhao2019pyod} for comparison with NSA techniques. Before applying NSA, we did t-distributed stochastic neighbourhood embedding (t-SNE \cite{van2008visualizing}) to reduce the dimensionality. We used t-SNE as it provided better visualization and reductions than PCA analysis as PCA works by rotating the vectors for preserving variance, which is suitable for linear, but t-SNE works by minimizing the distance between the point in a Gaussian, and it is a non-linear technique. Unlike PCA, it doesn't get influenced by outliers as it prioritizes local structure instead of the global structure. We applied V-detector methods to all datasets two-dimensional representation space, which acquire using the t-SNE method. We experimented ten times with different test-train split and provided the average of the experimented result.

\subsection{Our observations}
Results of Pentagon and Ring are illustrated in figure \ref{fig:pentagonresults1} and \ref{fig:pentagonresults} and in the table \ref{tab:results}. A visual comparison chart for performance and time are illustrated in the figure \ref{fig:resultcom} and \ref{fig:timecom}. 

From our experimental results (shown in the table \ref{tab:results}), we noticed that some methods outperform NSA-based approach in terms of accuracy and time consumption. But a close examination reveals that these methods performed better in pentagon dataset than for Ring dataset, where NSA based methods did better in Ring dataset,  indicating that for the complex nonlinear representation of data where pockets of an outlier can occur inside inlier or vice versa, that situation NSA can outperform other models. Among the NSA variations, V-detector has better performance. Linear model OCSVM has exhibited good performance in pentagon but performed poorly for Ring. This trend is also consistent with other linear model such as LMDD and MCD. As Ring was less linear than a Pentagon, performance decrease result is expected from linear models. It is also noticeable that OCSVM and MCD have similar execution times as NSA variations, but LMDD has significantly higher time consumption. For visual analysis, in figure  \ref{fig:vdet}, V-detector can detect but has slightly skewed in border areas. For the linear models in figure \ref{fig:pentagonresults} and \ref{fig:pentagonresults1}, we can see MCD and LMDD has big miss classification for a specific cluster area (more evident in Ring). OCSVM has small distributed packets of misclassifications around the border. 

Proximity-based methods LOF shows similar performance for both data. However, the connectivity-based outlier calculation failed to perform. HBOS and KNN did exceptionally well, and both were consistent for Pentagon and Ring both. The histogram-based outlier method did well will Pentagon but not so well for the Ring. Interesting was SOD, which performs better in Ring than  Pentagon, which is opposite trend than other models. In terms of execution time, COF is exceptionally higher than others. For visual analysis, for the RING, we can see HBOS has few large packets of miss classification holes, but in LOF and KNN, very tiny misclassification holes are distributed sparsely. For the pentagon, it was seen that HBOS is very skewed in the border areas where KNN and LOF have a less uneven distribution of misclassification holes.

In probability-based models, ABOD was consistent where COPD has shown a decline for the Ring dataset. ABOD outperform others in terms of performance, but it took a slightly higher execution time. The SOS method performed very poorly for our dataset, which shows the stochastic method's limitations; it also carried significantly higher execution time. ABOD has sparsely distributed holes in the visualization, but these are less than the SOS method.

For ensemble methods, Isolation forest has minimal decline from Pentagon to Ring, but feature bagging and LSCP are very consistent where LODA methods show a more considerable drop. However, all of these methods have a decent result (at least for the Pentagon). In terms of execution, LODA and FB outperform others, and LSCP had a much higher execution rate than higher. XGBOD method failed, but it was due to not having proper train data. We also did PCA representation, and it was evident that PCA was unable to perform for nonlinear data as it has shown the highest drop for Ring than Pentagon among all other models. But PCA has lowest execution time than others. LSCP and FB have similar representations in the visualization where both have sparsely distributed holes, but isolation forest holes were all clustered in the boundary region. Interesting was the LODA method where only a specific large cluster of holes was formed for the Ring and the outside border area for the Pentagon.

Both autoencoder based model shows a sharp decline for Ring, VAE did better perform that other AE, and it also took higher execution time. The limitations of this method are higher time consumption to train the model. GAN based model SOGAL and MOGAL were not able to perform, probably due to training failures. Based on their time consumption rate, it is evident that these methods are not applicable for this type of data representation.Both autoencoders have similar representation, and the noticeable thing is they have a single large miss classification cluster in a particular place. GAN based both model also has the same feature, but it misrepresented in testing time.

In the figure \ref{fig:comparisonfigse}, we presented 15 datasets in 2d representation space after using the t-SNE method for dimensionlity reduction. It is evident that some datasets are linearly separable (MUSK, OPTdigits), and others are hard to distinguish, such as Pima and Letters. We observed that some OCC has a lower variation of performance in all datasets than others. We consider these as good consistencies. In the figure \ref{fig:rr}, standard deviation indicates this consistency measures.

In the table \ref{tab:otherdatasets}, we presented the ROC performance of OCC techniques and NSA v-detector. The V-detector was consistent in its performance, where other methods did better in one dataset but  not so good for other data. We also noted that V-detector performance is similar to PCA and KNN techniques. One of the ensemble technique FB showed average performance, but another ensemble technique (IF) while outperformed other techniques and was most consistent. Also, the LOF method performed poorly, but clustering-based LOF improved the result. But these proximity-based methods are inconsistent considering all datasets. MCD outperforms OCSVM in performance consistency. On average, V-detector performs lower than Isolation forest and MCD, but it has better consistent performance except Isolation forest.  Figure \ref{fig:comparisonfigse}, showed that dataset PIMA has outlier and inlier are overlapped because of information loss in t-SNE dimensionality reduction, and this affected the performance of the V-detector for this dataset. Similarly, excellent performance of t-SNE in optdigits can be attributed to the t-SNE reductions. 

Figure \ref{fig:rr} illustrated that the isolation forest approach has best consistency in performance for all dataset and LOF has worst performance. The drop of V-detector methods consistency over the dataset were due to limitation of dimensionailty reduction technique, Our extended experiments with combination of PCA-transformed dataset and kernel-PCA\cite{hoffmann2007kernel} transformed data-set exhibited that this limitation can be addressed in NSA using V-detector.

\section{Challenges and Future Direction}

Our comparison analysis suggested that NSA alternate models' competitor can be autoencoder based model as they perform consistently as NSA. GAN-based models' current state is not adequate for the representation typically set NSA to perform well. Also, it is not suitable for all use cases. Probabilistic models ABOD also a good contender for NSA in terms of performance, but NSA can edge ABOD in execution time. LSCP is the only other consistent model and takes less time as it runs in parallel. If the NSA model can utilize a distributed system's power, it can be better time-efficient than LSCP.

Most of the NSA based researchers pick OCSVM and IF for the benchmark. Recent alternatives like autoencoder based or GAN based method are often ignored. We suggest researchers compare their result with contemporary alternate methods such as GAN based or autoencoder based algorithms for proper bench-marking. Some of these methods are too computationally expensive to implement, and some are vulnerable to high-level data. That's why it was essential to compare with multiple alternate methods. 

There is much research towards dynamic detector size and positions but not much research on self data representation spaces. New ideas on self data representations can open a new door for NSA. Much research on NSA started to use non-self data in the detector generation phase for better optimization. This feature limits the fundamental ability of the NSA as a one-class detector. Uses of a well-establish optimization algorithm can speed up the NSA detector generation process, but these optimization algorithms often need much data, which also backside for NSA applicability. NSA can be benefited by the recent improvement in the distributed bigdata system. There is little research to make NSA compliant with big data, so this remains an open challenge to other researchers. We also think NSA researchers neglect the computer vision domain, and there are enormous potential and space for work in computer vision with NSA. 

\section{Conclusion}
 This paper showed that the research of the Negative Selection Algorithm could be grouped into four eras, and each period has different research trends. Our study also provided a detailed discussion of the latest Negative Selection models and Negative Selection based applications. From our analysis, it is apparent that NSA operates better for nonlinear representation than most other techniques, and it can beat a neural-based model for time consummation. That is why research for further NSA's advancement to compete with recent models is promising.

\bibliographystyle{plain}
\bibliography{sample-base}

\end{document}